%% file: iclr2022_conference.tex
\documentclass{article} 

\usepackage{iclr2022_conference,times}

\input{math_commands.tex}

\usepackage{hyperref}
\usepackage{url}
\usepackage{booktabs}       
\usepackage{amsfonts}       
\usepackage{nicefrac}       
\usepackage{microtype}      
\usepackage{graphicx}
\usepackage{cancel}
\usepackage{bbm}
\usepackage{color}
\usepackage{multirow}
\usepackage{amsmath}
\usepackage{amsthm}
\usepackage{algorithm}
\usepackage{algpseudocode}
\usepackage{caption,subcaption}
\usepackage{chngcntr}
\usepackage{natbib}

\newtheorem{theorem}{Theorem}
\counterwithin*{theorem}{section} 
\newtheorem{prop}{Proposition}
\counterwithin*{prop}{section}

\newtheorem{remark}{Remark}

\title{Self-ensemble Adversarial Training for Improved Robustness}

\newcommand*{\affaddr}[1]{#1} 
\newcommand*{\affmark}[1][]{\textsuperscript{#1}}

\author{%
Hongjun Wang\affmark[1]\quad Yisen Wang\affmark[1,2]\footnotemark[2] \\
\affaddr{\affmark[1] Key Lab. of Machine Perception (MoE), School of Artificial Intelligence, Peking University}\\
\affaddr{\affmark[2] Institute for Artificial Intelligence, Peking University}
}

\iclrfinalcopy 
\begin{document}

\renewcommand{\thefootnote}{\fnsymbol{footnote}} 
\footnotetext[2]{Correspondence to: Yisen Wang (yisen.wang@pku.edu.cn)} 

\maketitle

\begin{abstract}
Due to numerous breakthroughs in real-world applications brought by machine intelligence, deep neural networks (DNNs) are widely employed in critical applications.
However, predictions of DNNs are easily manipulated with imperceptible adversarial perturbations, which impedes the further deployment of DNNs and may result in profound security and privacy implications.
By incorporating adversarial samples into the training data pool, adversarial training is the strongest principled strategy against various adversarial attacks among all sorts of defense methods. 
Recent works mainly focus on developing new loss functions or regularizers, attempting to find the unique optimal point in the weight space.
But none of them taps the potentials of classifiers obtained from standard adversarial training, especially states on the searching trajectory of training. 
In this work, we are dedicated to the weight states of models through the training process and devise a simple but powerful \emph{Self-Ensemble Adversarial Training} (SEAT) method for yielding a robust classifier by averaging weights of history models.
This considerably improves the robustness of the target model against several well known adversarial attacks, even merely utilizing the naive cross-entropy loss to supervise.
We also discuss the relationship between the ensemble of predictions from different adversarially trained models and the prediction of weight-ensembled models, as well as provide theoretical and empirical evidence that the proposed self-ensemble method provides a smoother loss landscape and better robustness than both individual models and the ensemble of predictions from different classifiers.
We further analyze a subtle but fatal issue in the general settings for the self-ensemble model, which causes the deterioration of the weight-ensembled method in the late phases\footnote{Code is available at \url{https://github.com/whj363636/Self-Ensemble-Adversarial-Training}}. 
\end{abstract}

\section{Introduction}
Deep learning techniques have showed promise in disciplines such as computer vision \citep{DBLP:conf/nips/KrizhevskySH12,DBLP:conf/cvpr/HeZRS16}, natural language processing \citep{DBLP:conf/nips/VaswaniSPUJGKP17,DBLP:conf/naacl/DevlinCLT19}, speech recognition \citep{DBLP:conf/interspeech/SakSRB15,wang2017residual} and even in protein structural modeling \citep{AlphaFold2020,AlphaFold2021}. 
However, even for those efforts surpassing human-level performance in several fields, 
deep learning based methods are vulnerable to adversarial examples generated by adding small perturbations to natural examples \citep{DBLP:journals/corr/SzegedyZSBEGF13,DBLP:journals/corr/GoodfellowSS14}.

Following the discovery of this adversarial vulnerability \citep{ma2019understanding,DBLP:conf/cvpr/0005WLZL20,niu2021moire}, numerous defense approaches for protecting DNNs from adversarial attacks have been proposed \citep{DBLP:conf/sp/PapernotM0JS16,DBLP:conf/ndss/Xu0Q18,DBLP:conf/cvpr/LiaoLDPH018,DBLP:conf/aaai/RossD18,bai2019hilbert,DBLP:conf/iclr/QinFSRCH20,wang2020hamiltonian,bai2021improving,huang2021exploring}. 
Because it is a straightforward and effective data-driven strategy, among these methods, adversarial training \citep{DBLP:conf/iclr/MadryMSTV18} has become the de-facto way for strengthening the robustness of neural networks.
The core of adversarial training is proactively inducing a robust model by forcing DNNs to learn adversarial examples generated from the current model, which helps DNNs gradually make up for their deficiencies.
Therefore, it is a practical learning framework to alleviate the impact of adversarial attacks.

As a saying goes, a person may go faster but a team can go farther. An ensemble of classifiers fused by output aggregates the collective wisdom of all participated classifiers, which outperform the decision made by a single classifier. Many ensemble methods \citep{DBLP:conf/iclr/TramerKPGBM18,DBLP:conf/icml/PangXDCZ19} have shown their abilities to boost adversarial robustness. 
Despite averaged predictions of different models enables the aggregation against a large number of attacks, it is still a heavy computing burden to adversarially train all the member models. 
What is worse, the architectures and parameters of \emph{every} member model, as well as their related contribution factors, should be safely saved and need performing forward propagation for every member repeatedly, which is a nightmare when dealing with massive components.
Despite the fact that the ensemble technique adopts a variety of architectures and parameters, these models may have similar decision-making boundaries. Therefore, it is natural to wonder:
\begin{quote}
    \emph{Could we develop the potential of the classifier itself to build a robust model rather than rely on multiple seemingly disparate classifiers? Could we make use of states on the searching trajectory of optimization when performing adversarial training?}
\end{quote}
Motivated by these questions, we discuss the relationship between an ensemble of different classifiers fused by predictions and by parameters, and present a computation-friendly self-ensemble method named SEAT.
In summary, this paper has the following contributions:
\begin{itemize}
  \item By rethinking the standard adversarial training and its variants, instead of developing a well-designed loss to find a single optimal in the weight space, we design an efficient algorithm called \emph{Self-Ensemble Adversarial Training} (SEAT) by uniting states of every history model on the optimization trajectory through the process of adversarial training. Unlike the traditional ensemble technique, which is extremely time-consuming and repeating adversarial training procedures for individuals, SEAT simply requires training once. 
  \item We present a theoretical explanation of the differences between the two ensemble approaches discussed above and visualize the change of loss landscape caused by SEAT. Furthermore, while using data augmentation of external data can experimentally fix the deterioration during the late stages \citep{gowal2020uncovering,rebuffi2021fixing}, we further investigate a subtle but fatal intrinsic issue in the general settings for the self-ensemble model, which causes the deterioration of the weight-ensembled method. 
  \item We thoroughly compare adversarial robustness of SEAT with other models trained by the state-of-the-art techniques against several attack methods on CIFAR-10 and CIFAR-100 datasets. Results have shown that the SEAT method itself can efficiently search in the weight space and significantly improve adversarial robustness. 
\end{itemize}

\section{Review: Standard Adversarial Training and its Successors}
In this section, we first introduce basic concepts of adversarial training. Then we briefly review some variants and enhancements of standard adversarial training.

\subsection{Preliminaries}
Consider a standard image task which involves classifying an input image into one of $C$ classes, let $x \in[0,1]^{n}$ denote the input natural image and $y \in\{1,2, \ldots, C\}$ be the ground truth label, sampled from a distribution $\mathbb{D}$. For adversarial attack, the goal of the adversary is to seek a malignant example in the vicinity of $x$, by adding a human imperceptible perturbation $\varepsilon \in \mathbb{R}^{n}$.
To exactly describe the meaning of ``imperceptible", this neighbor region $\mathbb{B}_{\varepsilon}(x)$ anchored at $x$ with apothem $\varepsilon$ can be defined as $\mathbb{B}_{\varepsilon}(x)=\left\{(x^{\prime},y) \in \mathcal{D} \mid \left\|x-x^{\prime}\right\| \leq \varepsilon\right\}$.
We also define a loss function: $\ell\overset{def}{=}\Theta\times\mathbb{D}\rightarrow [0, \infty)$, which is usually positive, bounded, and upper-semi continuous \citep{DBLP:journals/mor/BlanchetM19,Villani2003TopicsIO,DBLP:conf/colt/BartlettM01} for all $\boldsymbol\theta\in\Theta$. 
Instead of the original formulation of adversarial training \citep{DBLP:journals/corr/SzegedyZSBEGF13,DBLP:journals/corr/GoodfellowSS14}, which mixes natural examples $x$ and adversarial examples $x^{\prime}$ together to train models, the adversarial risk minimization problem can be formulated as a two-player game that writes as below:
\begin{equation}
\label{eqn:AT_basic}
\underbrace{\inf _{\boldsymbol\theta \in \Theta} \mathbb{E}_{(x, y) \sim \mathbb{D}}}_{\hbox{\footnotesize{Outer minimization}}}\underbrace{\sup _{x^{\prime} \in \mathbb{B}_{\varepsilon}(x)} \ell\left(\boldsymbol\theta;x^{\prime}, y\right)}_{\hbox{\footnotesize{Inner maximization}}}.
\end{equation}
In general, the adversary player computes more and more harmful perturbations step by step, where the PGD method \citep{DBLP:conf/iclr/MadryMSTV18} is commonly applied:
\begin{equation}
x^{t+1}=\Pi_{\mathbb{B}_{\varepsilon}(x)}\left(x^{t}+\kappa\operatorname{sign}\left(\nabla_{x} \ell(x, y)\right)\right),
\end{equation}
where $\kappa$ is the step size and the generated samples at each iteration will be projected back to the $\mathbb{B}_{\varepsilon}(x)$.
When finding $x^{\prime}$ at their current optimal in the inner maximization, the model parameters will be updated according to these fixed samples only in the outer minimization. 

\subsection{Related Work}
Emerging adversarial training methods are primarily concerned with strengthening their robustness through empirical defense. Here we roughly separate these improvements into two categories: (1) \emph{sample-oriented} (i.e. for $x^{\prime}$ in the inner maximization) and (2) \emph{loss-oriented} (i.e. for $\ell$ in the outer minimization) improvement.

From the side of the former, \cite{DBLP:journals/corr/BalujaF17} utilize a GAN framework to create adversarial examples via the generative model. \cite{wang2019dynamic} proposes a criterion to control the strength of the generated adversarial examples while FAT \citep{DBLP:conf/icml/ZhangXH0CSK20} control the length of searching trajectories of PGD through the lens of geometry.

For the latter, a stream of works have been carried out with the goal of establishing a new supervised loss for better optimization. Usually, the loss $\ell$ in the both inner maximization and outer minimization\footnote{For simplification, here we ignore the implemented difference of loss in the inner maximization and the outer minimization (e.g. TRADES and MMA). The reference of $\ell$ can be contextually inferred easily.} are cross-entropy defined as: $\ell_{ce}=-\boldsymbol{y}^{T} \log \boldsymbol{q}$ when given the probability vector $\boldsymbol{q}=[\mathrm{e}^{f_{\boldsymbol\theta}(x^{\prime})_{1}}, \cdots, \mathrm{e}^{f_{\boldsymbol\theta}(x^{\prime})_{c}}] / \sum_{k=1}^{c} \mathrm{e}^{f_{\boldsymbol\theta}(x^{\prime})_{k}}$. The addition of a regularizer to the vanilla loss is one branch of the modification. Representative works like ALP \citep{DBLP:journals/corr/abs-1803-06373}, TRADES \citep{DBLP:conf/icml/ZhangYJXGJ19}, and VAT \citep{DBLP:journals/pami/MiyatoMKI19} introduce a regularization term to smooth the gap between the probability output of natural examples and adversarial ones, which can be denoted as: 
\begin{equation}
\label{eqn:regularizer}
\ell_{reg}=\ell_{ce}+\eta\sum_{(x_i, y_i)\sim \mathcal{B}}\mathcal{R}(p(\boldsymbol\theta;x^{\prime}_i, y_i), p(\boldsymbol\theta;x_i, y_i)),
\end{equation}
where $\eta$ is the regularization hyper-parameter and $\mathcal{R}(\cdot, \cdot)$ is $l_2$ distance for ALP and VAT\footnote{For VAT, $\theta$ in the first term and the second term are different.}, and Kullback-Leibler divergence for TRADES. $\mathcal{B}$ stands for the sampled minibatch.
Another branch of the loss modification is to reweight the contributions of each instance in the minibatch based on their intrinsic characteristics. This can be formulated by:
\begin{equation}
\label{eqn:reweighting}
\ell_{rew}=\sum_{(x_i, y_i)\sim \mathcal{B}} w(\boldsymbol\theta;x_{i}, x^{\prime}_i, y_{i}) \ell_{ce}\left(\boldsymbol\theta;x_i, x^{\prime}_i, y_i\right),
\end{equation}
where $w(\cdot)$ is a real-valued function mapping the input instances to an importance score. MMA \citep{DBLP:conf/iclr/DingSLH20} separately considers the correct and incorrect classification on natural cases and switches between them by learning a hard importance weight:

\begin{equation}
\label{eqn:AT_mma}
\ell_{\operatorname{MMA}} = \sum_{(x_i, y_i)\sim \mathcal{B}} w_1(\boldsymbol\theta;x_{i}, y_{i}) \ell_{ce}\left(\boldsymbol\theta;x_i, y_i\right) + w_2(\boldsymbol\theta;x_{i}, y_{i}) \ell_{ce}\left(\boldsymbol\theta;x^{\prime}_i, y_i\right),
\end{equation}
where $w_1$ and $w_2$ are the indicator function $\mathbbm{1}\left(f_{\boldsymbol{\theta}}\left(x_{i}\right) \neq y_{i}\right)$ and $\mathbbm{1}\left(f_{\boldsymbol{\theta}}\left(x_{i}\right)=y_{i}\right)$, respectively. Inspired by the geometry concern of FAT, GAIRAT \citep{DBLP:conf/iclr/ZhangZ00SK21} evaluates $w(\boldsymbol\theta;x, y)=(1+\tanh (c_1 - 10c_2 + 5))/2$ to adaptively control the contribution of different instances by their corresponding geometric distance of PGD, where $c_1$ is a hyper-parameter and $c_2$ is the ratio of the minimum successful iteration numbers of PGD to the maximum iteration step. Embracing both two schools of improvement, MART \citep{DBLP:conf/iclr/0001ZY0MG20} adds a new regularization term apart from KL divergence and explicitly assigns weights within regularizers:
\begin{equation}
\label{eqn:AT_mart}
\ell_{\operatorname{MART}} = \ell_{ce} + \sum_{(x_i, y_i)\sim \mathcal{B}} w(\boldsymbol\theta;x_{i}, y_{i})\mathcal{R}_{KL}(p(\boldsymbol\theta;x^{\prime}_i, y_i), p(\boldsymbol\theta;x_i, y_i)) + \mathcal{R}_{mag}(p(\boldsymbol\theta;x^{\prime}_i, y_i)),
\end{equation}
where $\mathcal{R}_{KL}$ is Kullback-Leibler divergence same as TRADES and $w_i=\left(1-p(\boldsymbol{\theta};x_i,y_i)\right)$ is a softer scheme when compared with MMA. The margin term $\mathcal{R}_{mag}=-\log \left(1-\max _{k \neq y_{i}} p_{k}\left(\boldsymbol{\theta};x_{i}^{\prime},y_i\right)\right)$ aims at improving the decision boundary. 

The above methods endeavor to solve the highly non-convex and non-concave optimization in Eqn \ref{eqn:AT_basic}. However, they assume that the optimal weights of a classifier appear in its prime and abandon all the states along the route of optimization, which are beneficial to approach the optimal.

\section{Methodology}
In this section, we first discuss the traditional ensemble methods based on predictions of several individual classifiers. Then, we propose a Self-ensemble Adversarial Training (SEAT) strategy fusing weights of individual models during different periods, which is both intrinsically and computationally convenient. Following that, we further provide theoretical and empirical analysis on why the prediction of such an algorithm can make a great difference from that of simple ensembling classifiers.

\subsection{Prediction-oriented Ensemble}
Ensemble methods are usually effective to enhance the performance \citep{DBLP:conf/icml/CaruanaNCK04,DBLP:conf/nips/GaripovIPVW18} and improve the robustness \citep{DBLP:conf/iclr/TramerKPGBM18,DBLP:conf/icml/PangXDCZ19}. 
For clarity, we denote $\mathcal{F}$ as a pool of candidate models where $\mathcal{F}=\left\{f_{\boldsymbol\theta_1},\cdots, f_{\boldsymbol\theta_n}\right\}$. To represent the scalar output of the averaged prediction of candidate classifiers over $C$ categories, we define the averaged prediction $\bar{f}$ involving all candidates in the pool $\mathcal{F}$ as:
\begin{equation}
\begin{aligned}
\label{eqn:ave_output}
\bar{f}_{\mathcal{F}}(x,y)&=\sum_{i=1}^{n} \beta_{i} f_{\boldsymbol\theta_i}(x,y) \\
s.t. \  &\sum_{i=1}^{n} \beta_{i}=1,
\end{aligned}
\end{equation}
where $\beta$ represents the normalized contribution score allotted to the candidates. It is logical to assume that $\forall \beta_{i}>0$ since each candidate model must be well-trained, otherwise it will be expelled from the list. Note that here we only discuss whether the participated model performs well on its own, not whether it has a positive impact on the collective decision making, which is a belated action.

\begin{algorithm}[!t]
\small
   \caption{Self-Ensemble Adversarial Training (SEAT)}
   \label{alg:SEAT}
\begin{algorithmic}
   \State {\bfseries Input:} A DNN classifier $f_{\boldsymbol\theta}(\cdot)$ with initial learnable parameters $\boldsymbol\theta_0$ and loss function $\ell$; data distribution $\mathbb{D}$; number of iterations $N$; number of adversarial attack steps $K$; magnitude of perturbation $\varepsilon$; step size $\kappa$; learning rate $\tau$; exponential decay rates for ensembling $\alpha$; constant factor $c$.
   \State Initialize $\boldsymbol\theta \leftarrow \boldsymbol\theta_0$, $\tilde{\boldsymbol\theta} \leftarrow \boldsymbol\theta$.
   \For {t $ \leftarrow 1, 2, \cdots , N$}
    \State Sample a minibatch (x, y) from data distribution $\mathbb{D}$
    \State $x^{\prime}_{0} \leftarrow x+\varepsilon$, $\varepsilon\sim \operatorname{Uniform}(-\varepsilon,\varepsilon)$.
    \For {k $ \leftarrow 1, 2, \cdots , K$}
      \State $x^{\prime}_{k} \leftarrow \Pi_{x^{\prime}_{k} \in \mathbb{B}_{\varepsilon}(x)}\left(\kappa \operatorname{sign}\left(x^{\prime}_{k-1}+\nabla_{x^{\prime}_{k-1}} \ell(\boldsymbol\theta; (x^{\prime}_{k}, y))\right)\right)$
    \EndFor
    \State $\boldsymbol{g}_{\boldsymbol\theta_t} \leftarrow \mathbb{E}_{(x, y)}\left[\nabla_{\boldsymbol\theta_t} \ell(\boldsymbol\theta_t;(x^{\prime}_k, y))\right]$ in every minibatch
    \State Calculate $\tau_t$ according to the current iterations
    \State $\boldsymbol\theta_t \leftarrow \boldsymbol\theta_t-\tau_t \boldsymbol{g}_{\boldsymbol\theta_t}$
    \State $\alpha^{\prime} \leftarrow \operatorname{min}\left(\alpha, \frac{t}{t+c}\right)$
    \State $\tilde{\boldsymbol\theta} \leftarrow \alpha^{\prime}\tilde{\boldsymbol\theta} + (1-\alpha^{\prime})\boldsymbol\theta_t$
   \EndFor
   \State \textbf{Return} A self-ensemble adversarial training model $f_{\tilde{\boldsymbol\theta}}(\cdot)$
\end{algorithmic}
\end{algorithm}

\subsection{Self-ensemble Adversarial Training}
In order to average weights of a model, it is natural to choose an analogous form as Eqn \ref{eqn:ave_output} to obtain the predicted weights: $\tilde{\boldsymbol\theta}=\sum_{t=1}^{T} \beta_{t}\boldsymbol\theta_t, s.t. \sum_{t=1}^{T} \beta_{t}=1$.
However, such a simple moving average cannot keenly capture the latest change, lagging behind the latest states by half the sample width. 
To address this problem, we calculate the optimal weights by characterizing the trajectory of the weights state as an exponential moving average (EMA) for measuring trend directions over a period of time.
Intuitively, EMA not only utilizes recent proposals from the current SGD optimization but also maintains some influence from previous weight states:
\begin{equation}
\begin{aligned}
\tilde{\boldsymbol\theta}_T &=\alpha\tilde{\boldsymbol\theta}_{T-1} + (1-\alpha)\boldsymbol\theta_T \\
&=\sum_{t=1}^{T}(1-\alpha)^{1-\delta(t-1)}\alpha^{T-t}\boldsymbol\theta_t, 
\end{aligned}
\end{equation}
where $\delta(\cdot)$ is the unit impulse function (i.e., $\delta(0) = 1$, otherwise it is 0). 
Algorithm\ref{alg:SEAT} summarizes the full algorithm. 

In consideration of both using the moving average technique, we wonder whether the averaged prediction has a link to the prediction of a weight-averaged model. In fact, they are indeed related to each other to some extent:  
\begin{prop}
\label{pro:1}
(Proof in Appendix B) Let $f_{\boldsymbol\theta}\left(\cdot\right)$ denote the predictions of a neural network parametrized by weights $\boldsymbol\theta$. Assuming that $\forall \boldsymbol\theta \in \Theta$, $f_{\boldsymbol\theta}\left(\cdot\right)$ is continuous and $\forall (x, y)\in \mathbb{D}$, $f_{\boldsymbol\theta}(x, y)$ is at least twice differentiable. Consider two points $\boldsymbol\theta_t, \tilde{\boldsymbol\theta} \in\Theta$ in the weight space and let $\boldsymbol\xi=\boldsymbol\theta_t-\tilde{\boldsymbol\theta}$, for $t\in \left\{1,2,\cdots,T\right\}$, the difference between $\bar{f}_{\mathcal{F}}(x,y)$ and $f_{\tilde{\boldsymbol\theta}}(x, y)$ is of the second order of smallness if and only if $\sum_{t=1}^{T} (\beta_{t}\boldsymbol\xi^\top)=\boldsymbol 0$.
\end{prop}

\begin{figure}[!ht]
    \begin{minipage}[t]{.45\textwidth}
        \centering
        \includegraphics[width=\textwidth]{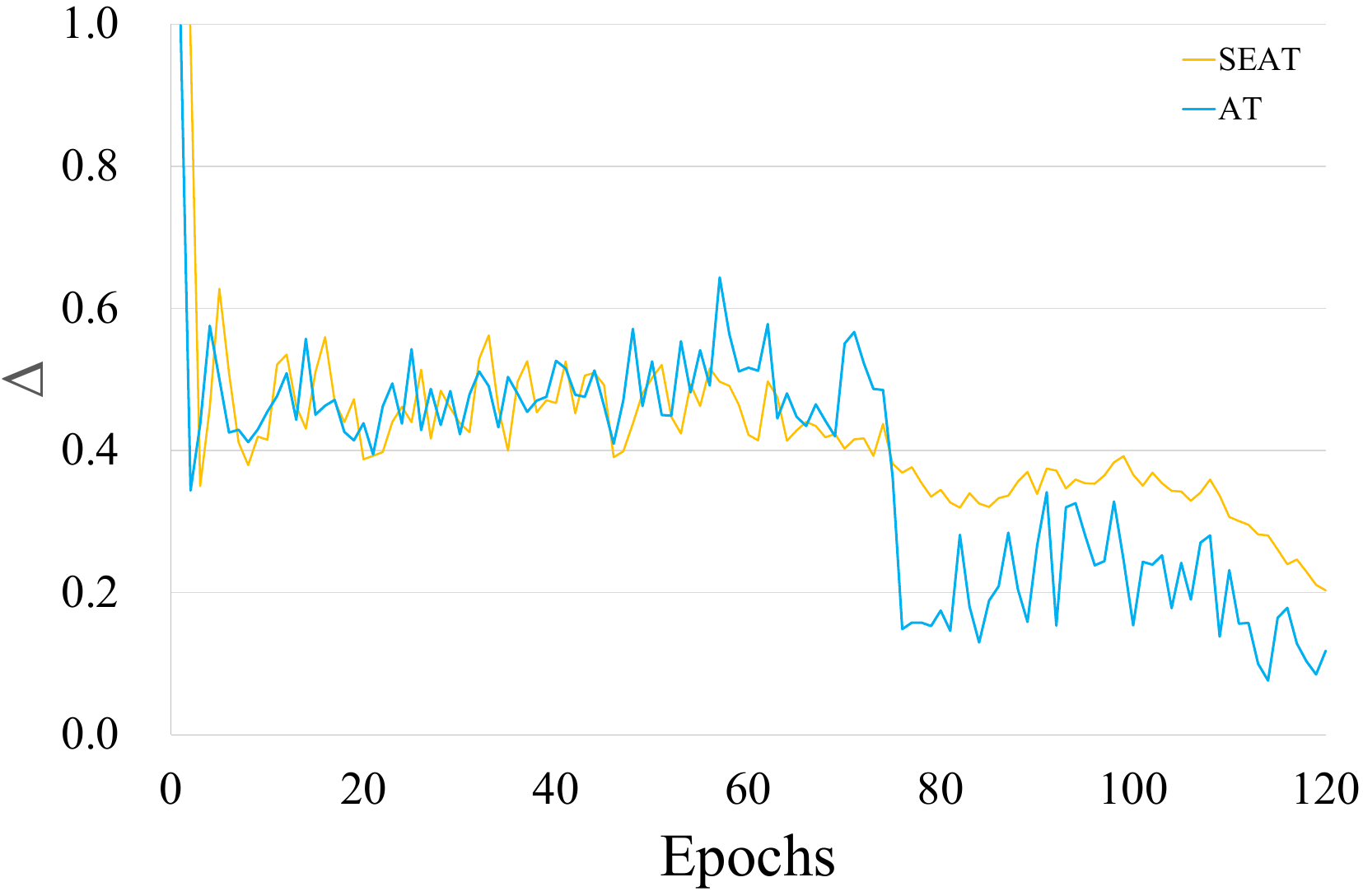}
        \subcaption{The effect of homogenization of models.}\label{fig:homogenization}
    \end{minipage}
    \hfill
    \begin{minipage}[t]{.45\textwidth}
        \centering
        \includegraphics[width=\textwidth]{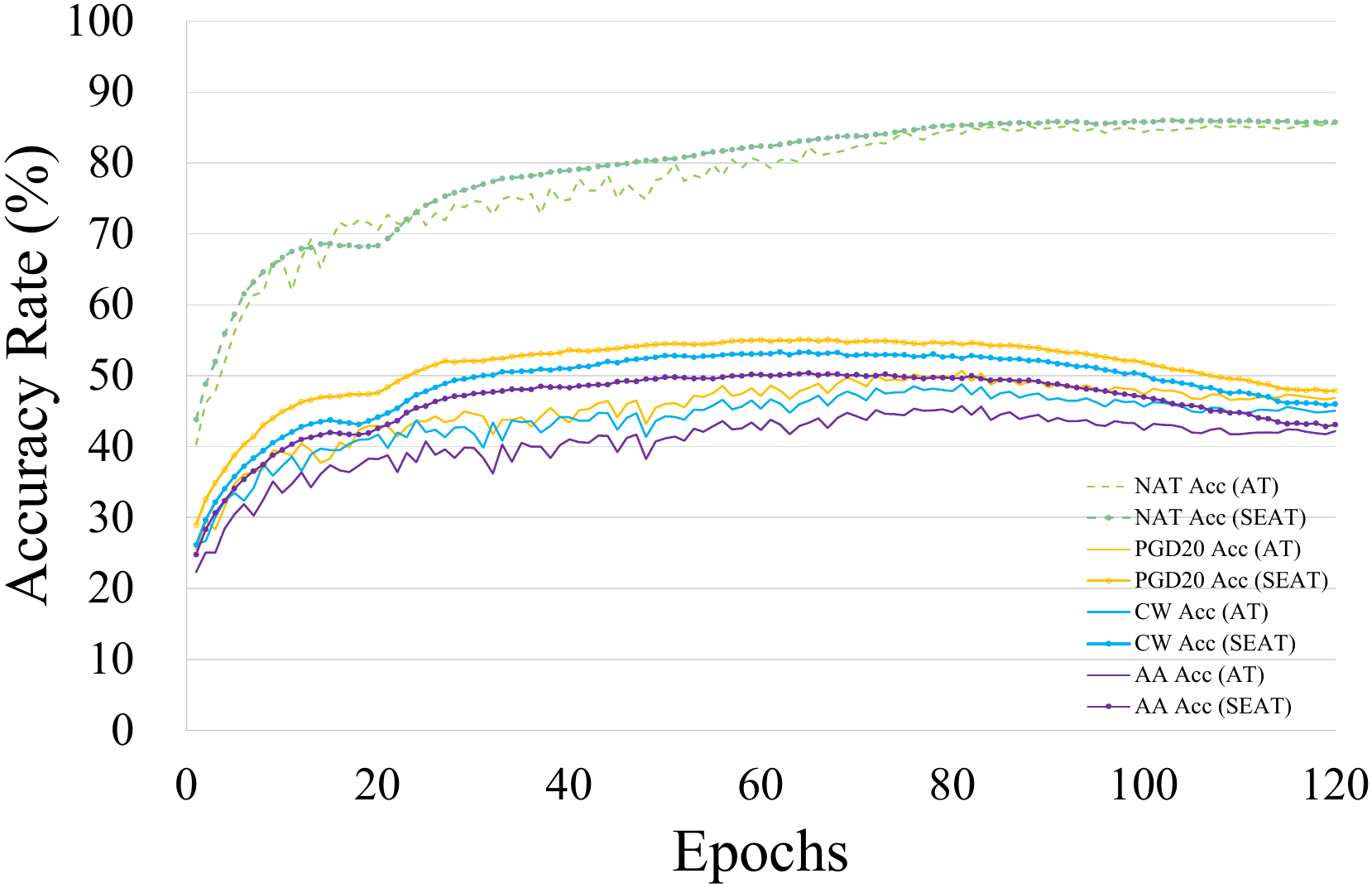}
        \subcaption{Robust accuracy of parameter-averaged ResNet18 against various attacks on CIFAR-10.}\label{fig:all_attack_resnet18}
    \end{minipage}
    \vspace{-0.1 in}
    \caption{The overall analyse of AT and SEAT.}\label{fig:first}
\end{figure}

Based on Proposition \ref{pro:1}, we could immediately obtain the below conclusions:
\begin{remark}
Note that it always constructs ensembles of the well-matched enhanced networks to obtain stronger defense, so it evenly assigns $\beta_1=\beta_2=\cdots=\beta_n=1/n$. However, things change in the self-ensemble setting since models obtained at relatively late stages will be much robust than the beginning. Based on this assumption, it has a non-decreasing sequence $\beta_1\leq \beta_2\leq \cdots\leq \beta_n, s.t. \sum_{i=1}^n\beta_i=1$. The inequality is tight only when the initial weight reaches its fixed point.
\end{remark}
In this case, the averaged prediction will predispose to the models obtained at the end of optimization and such a predisposition loses the superiority of ensemble due to the phenomenon of homogenization in the late phases. To provide an empirical evidence of homogenization of models at latter periods, we visualize the effect of homogenization along the training process in Figure \ref{fig:homogenization}. We define the homogenization of a model as: $\triangle_e=\frac{1}{||\mathcal{D}||}\sum\operatorname{min}_{i\in[1,m]}|f_{\boldsymbol\theta_e}(x,y)-f_{\boldsymbol\theta_{e-i}}(x,y)|$, used for calculating the difference of the output of history models over a period of time $m$. Note that $\triangle_e$ becomes smaller and smaller along with epochs passing, which proves the existence of homogenization.

Here we formally state the gap between such an averaged prediction and the prediction given by SEAT in the following theorem:
\begin{theorem}
\label{thm:1}
(Proof in Appendix B) Assuming that for $i,j\in \left\{1,\cdots,T\right\}$, $\boldsymbol\theta_i=\boldsymbol\theta_j$ if and only if $i=j$. The difference between the averaged prediction of multiple networks and the prediction of SEAT is of the second order of smallness if and only if $\beta_{i}=(1-\alpha)^{1-\delta(i-1)} \alpha^{T-i}$ for $i\in\left\{1,2,\cdots,T\right\}$.
\end{theorem}

Theorem \ref{thm:1} tells us that it is much harder for SEAT to approximate the averaged prediction of history networks than the EMA method. So the prediction of SEAT keeps away from the averaged prediction of models, which suffers from homogenization. To ensure whether the difference in Theorem \ref{thm:1} is benign, we provide empirical evidence of the improvement of the self-ensemble technique on the loss landscape. 

\begin{figure}[!t]
    \begin{minipage}[t]{.32\textwidth}
        \centering
        \includegraphics[width=\textwidth]{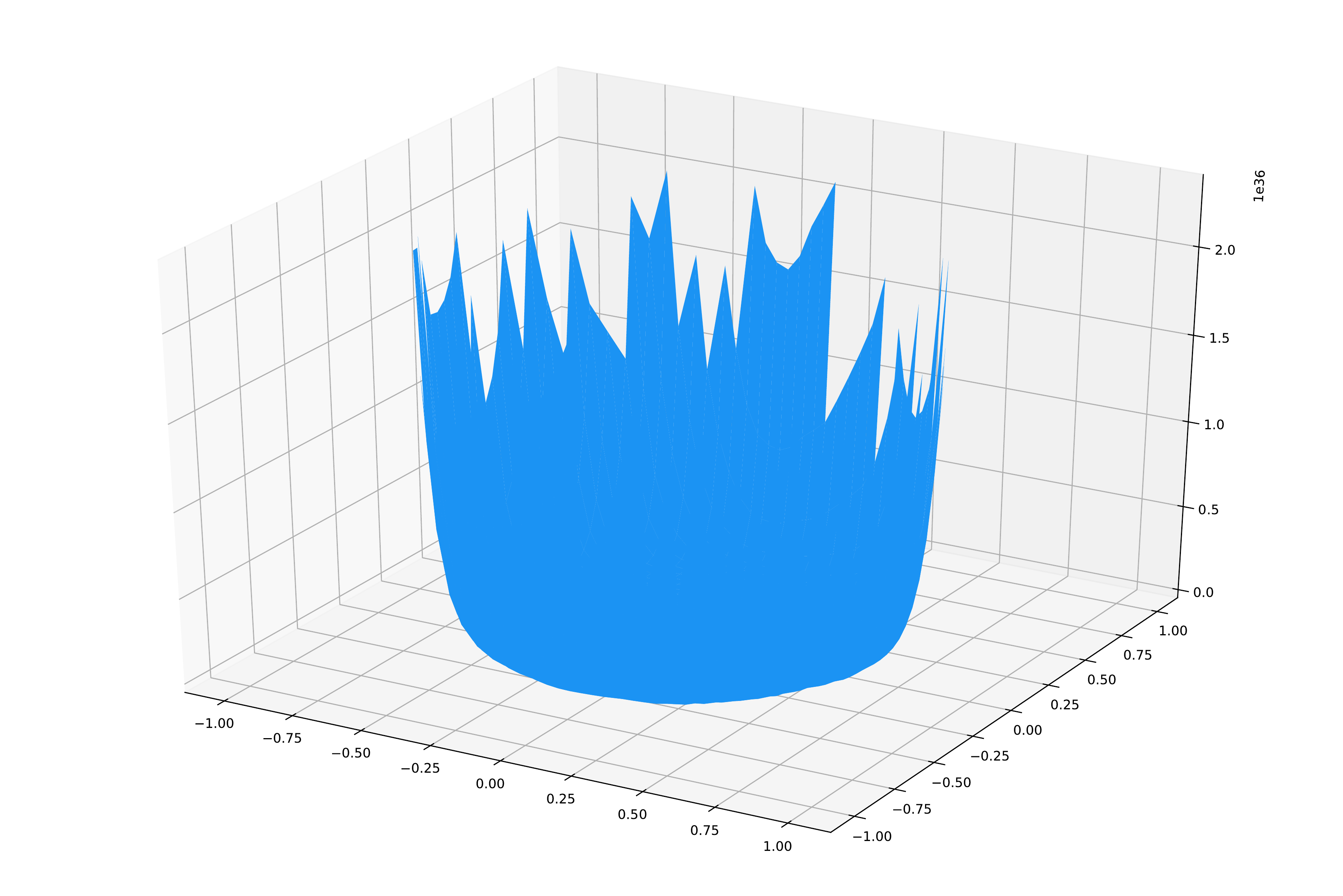}
        \subcaption{The best individual}\label{fig:1}
    \end{minipage}
    \hfill
    \begin{minipage}[t]{.32\textwidth}
        \centering
        \includegraphics[width=\textwidth]{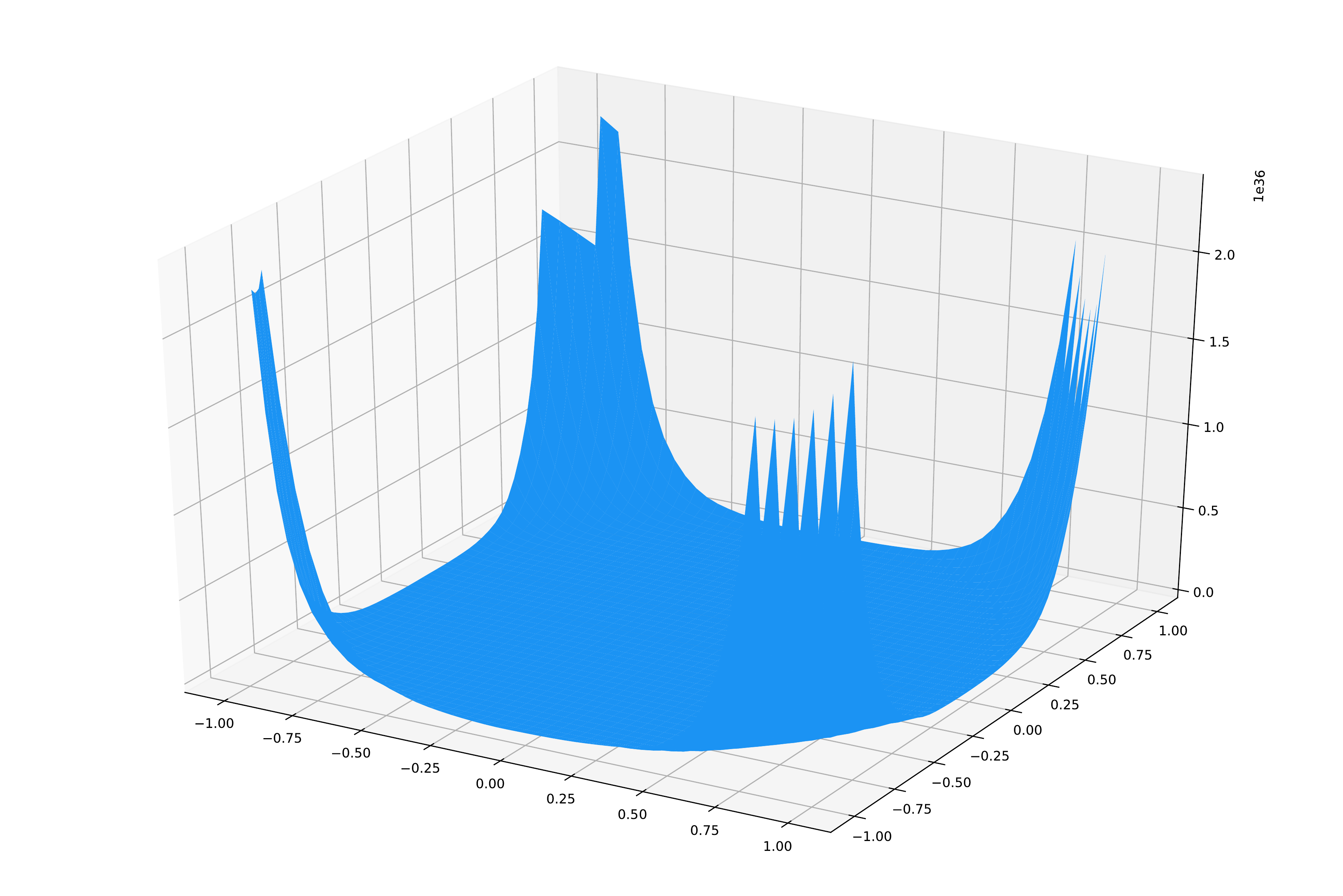}
        \subcaption{SEAT}\label{fig:2}
    \end{minipage}
    \hfill
    \begin{minipage}[t]{.32\textwidth}
        \centering
        \includegraphics[width=\textwidth]{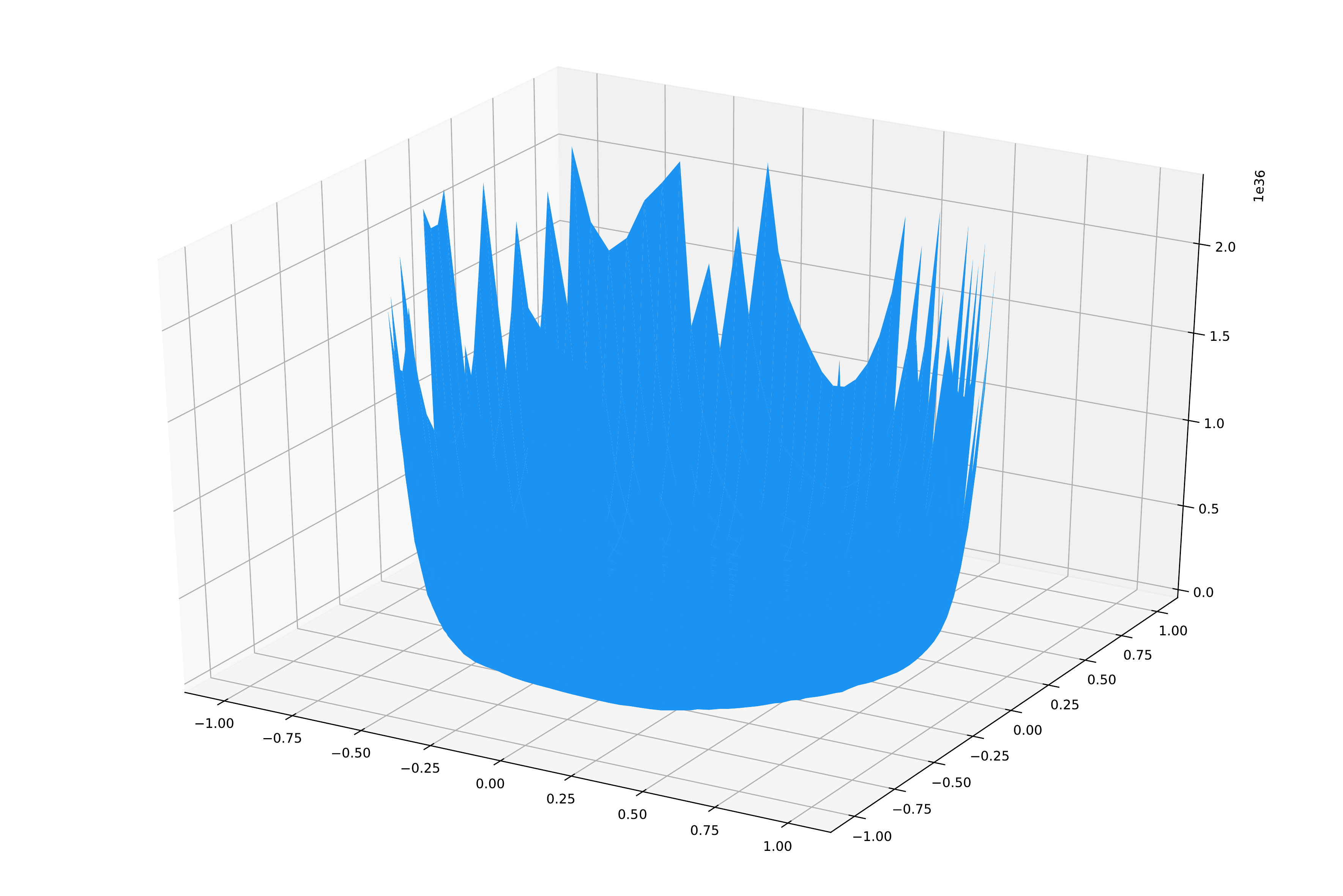}
        \subcaption{Deteriorated SEAT}\label{fig:3}
    \end{minipage}  
    \vspace{-0.1 in}
    \caption{The loss surfaces of ResNet18 on CIFAR-10 test data. The loss surface of the standard adversarially trained models transitions up and down, which means the loss of such a model changes dramatically as moving along some directions. In contrast, SEAT has a fairly smooth landscape.}\label{fig:loss_landscape}
    \vspace{-0.1 in}
\end{figure}

It is difficult for the traditional 1D linear interpolation method to visualize non-convexities \citep{DBLP:journals/corr/GoodfellowV14} and biases caused by invariance symmetries (e.g. batch normalization) in the DNN. \cite{visualloss,wu2020adversarial} address these challenges, providing a scheme called filter normalization to explore the sharpness/flatness of DNNs on the loss landscape. Let $\boldsymbol v_1$ and $\boldsymbol v_2$ are two random direction vectors sampled from a random Gaussian distribution. So we plot the value of loss function around $\theta$ when inputting data samples:
\begin{equation}
L(\boldsymbol\theta; \boldsymbol v_1, \boldsymbol v_2)=\ell\left(\boldsymbol\theta+m \boldsymbol v_1+n \boldsymbol v_2;x, y\right),
\end{equation}
where $m=\frac{\left\|\boldsymbol\theta\right\|}{\left\|\boldsymbol v_1\right\|}$, $n=\frac{\left\|\boldsymbol\theta\right\|}{\left\|\boldsymbol v_2\right\|}$, and $\left\|\cdot\right\|$ denotes the Frobenius norm.
Specifically, we apply the method to plot the surface of DNN within the 3D projected space. We visualize the loss values of ResNet18 on the testing dataset of CIFAR-10 in Figure \ref{fig:loss_landscape}. Each grid point in the x-axis represents a sampled gradient direction tuple $(\boldsymbol v_1,\boldsymbol v_2)$. It is clear that SEAT has a smooth and slowly varying loss function while the standard adversarially trained model visually has a sharper minimum with higher test error.

\textbf{Safeguard for the Initial Stage}: During the initial training period, the history models are less instrumental since they are not adequately learned. Therefore, it is not wise to directly make use of the ensembled weights, which may accumulate some naive weight states and do harm to approach the oracle. Therefore, we apply the protection mechanism for novice models at the beginning, denoted as $\alpha^{\prime}=\operatorname{min}\left(\alpha, \frac{i}{i+c}\right)$, where $c$ is a constant factor controlling the length of warm-up periods for history models.

\subsection{Deterioration: The Devil in the Learning Rate}
Having provided both theoretical and experimental support as well as a clear procedure of SEAT algorithm, we may take it for granted that this self-ensemble method could be integrated into the routine working flow as an off-shelf technique. But it is not the case.

As shown in Figure \ref{fig:2} and \ref{fig:3}, while the two self-ensemble models have \emph{almost} nearly identical settings, their loss landscapes are completely different (quantitative results can be referred to Appendix A.4). Despite the fact that the landscape of deteriorated SEAT is a little mild than the standard adversarially trained model, it is still extremely steep when compared with SEAT. This phenomenon about deterioration is also discovered by \cite{rebuffi2021fixing}. They combine model weight averaging with data augmentation and external data to handle the deterioration. The intention behind such a solution is robust overfitting \citep{DBLP:conf/icml/RiceWK20} of individuals. However, it is unclear why an ensemble model assembled from overfitting individuals performs poorly, while inferior models can still be aggregated to produce a superior one in the middle of training stages. In the rest of this section, we provide another solution from a quite new perspective: \emph{the strategy of learning rate}. 

\begin{prop}
\label{pro:2}
(Proof in Appendix B) Assuming that every candidate classifier is updated by SGD-like strategy, meaning $\boldsymbol\theta_{t+1} = \boldsymbol\theta_t -\tau_t \nabla_{\boldsymbol\theta_t}f_{\boldsymbol\theta_t}(x^{\prime}, y)$ with $\tau_1\geq\tau_2\geq\cdots\geq\tau_T>0$, the performance of self-ensemble model depends on learning rate schedules.
\end{prop}
As illustrated in Figure \ref{fig:homogenization}, we note that the phenomenon of homogenization exacerbates after each tipping point (e.g. Epoch 76, 91 and 111) when using the staircase strategy of learning rate. Combined with the trend of robust accuracy in Figure \ref{fig:all_attack_resnet18}, we find that 
the self-ensemble model cannot gain any benefits when the learning rate decays below a certain value. More results on different learning rate schedules can be found in Figure \ref{fig:lr}.

\section{Experimental Results}
In this section, we first conduct a set of experiments to verify the advantages of the proposed method. Then, we investigate how each component of SEAT functions and how it affects robust accuracy. 

\subsection{Experimental setup}
We mainly use ResNet18 and WRN-32-10 \citep{DBLP:journals/corr/ZagoruykoK16} for the experiments on CIFAR-10/CIFAR-100 and all images are normalized into $[0, 1]$. 
For simplicity, we only report the results based on $L_{\infty}$ norm for the non-targeted attack. We train ResNet18 using SGD with 0.9 momentum for 120 epochs and the weight decay factor is set to $3.5e^{-3}$ for ResNet18 and $7e^{-4}$ for WRN-32-10. For SEAT, we use the piecewise linear learning rate schedule instead of the staircase one based on Proposition \ref{pro:2}. The initial learning rate for ResNet18 is set to 0.01 and 0.1 for WRN-32-10 till Epoch 40 and then linearly reduced to 0.001, 0.0001 and 0.01, 0.001 at Epoch 60 and 120, respectively. The magnitude of maximum perturbation at each pixel is $\varepsilon=8/255$ with step size $\kappa=2/255$ and the PGD steps number in the inner maximization is 10. 
To evaluate adversarial defense methods, we apply several adversarial attacks including PGD \citep{DBLP:conf/iclr/MadryMSTV18}, MIM \citep{DBLP:conf/cvpr/DongLPS0HL18}, CW \citep{DBLP:conf/sp/Carlini017} and AutoAttack (AA) \citep{DBLP:conf/icml/Croce020a}. We mainly compare the following defense methods in our experiments:
\begin{itemize}
  \item \textbf{TRADES, MART, FAT and GAIRAT:} For TRADES and MART, we follow the official implementation of MART\footnote{https://github.com/YisenWang/MART} to train both two models for better robustness. The hyper-parameter $\eta$ of both TRADES and MART is set to 6.0. The learning rate is divided by 10 at Epoch 75, 90, 100 for ResNet18 and at 75, 90, 110 for WRN-32-10, respectively. For FAT and GAIRAT, we completely comply with the official implementation\footnote{https://github.com/zjfheart/Geometry-aware-Instance-reweighted-Adversarial-Training} to report results.
  \item \textbf{PoE:} We also perform the experiments on the Prediction-oriented Ensemble method (PoE). We select the above four advanced models as the candidate models. When averaging the output of these models, we assign different weights to individuals according to their robust accuracy. Specifically we respectively fix $\beta_{1,2,3,4}=0.1,0.2,0.3,0.4$ for FAT, GAIRAT, TRADES, and MART, named PoE. Considering the poor robust performance of FAT and GAIRAT, we also provide the PoE result by averaging the outputs of TRADES trained with $\lambda=1,2,4,6,8$ and assign the same $\beta$ to them, named PoE (TRADES). 
  \item \textbf{CutMix (with WA):} Based on the experimental results of \cite{rebuffi2021fixing}, Cutmix with a fixed window size achieves the best robust accuracy. We follow this setting and set the window size to 20. Since we do not have matched computational resources with Deepmind, we only train models for 120 epochs with a batch size of 128 rather than 400 epochs with a batch size of 512. Besides, we neither use Swish/SiLU activation functions \citep{DBLP:journals/corr/HendrycksG16} nor introduce external data and samples crafted by generative models \citep{DBLP:conf/iclr/BrockDS19,DBLP:conf/nips/HoJA20,DBLP:conf/iclr/Child21} for fair comparison.
\end{itemize}

\subsection{Robustness Evaluation of SEAT}
In this section, we fully verify the effectiveness of the SEAT method. We report both average accuracy rates and standard deviations. All results in Tables \ref{tab:Resnet18} and \ref{tab:WRN} are computed with 5 individual trials. Results on ResNet18 are summarized in Table \ref{tab:Resnet18}. Here, we mainly report the results on the CIFAR-10 dataset due to the space limitation. For the results on CIFAR-100, please refer to Appendix A.2.
From the table, we can see that the superiority of SEAT is apparent especially considering we only use candidate models supervised by the traditional cross-entropy loss without external data. 
Compared with two advanced loss-oriented methods (TRADES and MART), the results of SEAT are at least $\sim2\%$ better than TRADES and MART against CW and AA. 
We further emphasize that although the boost of SEAT regarding PGD attacks is slight when compared with GAIRAT, SEAT achieves startling robust accuracy results when facing CW and AA attacks. 
Note that AA is an ensemble of various advanced attacks (the performance under each component of AA can be found in Appendix A.1) and thus the robust accuracy of AA reliably reflects the adversarial robustness, which indicates that the robustness improvement of SEAT is not biased. 
\renewcommand{\arraystretch}{0.9}
\begin{table*}[!t]
\caption{Comparison of our algorithm with different defense methods using ResNet18 on CIFAR-10. The maximum perturbation is $\varepsilon=8/255$. Average accuracy rates (in \%) and standard deviations have shown that the proposed SEAT method greatly improves the robustness of the model.}\label{tab:Resnet18}
\centering
\footnotesize
\begin{tabular}{l|cccccc}
Method        & NAT            & $\operatorname{PGD}^{20}$          & $\operatorname{PGD}^{100}$         & MIM            & CW             & AA            \\ \hline\hline
AT            & 84.32$\pm$0.23         & 48.29$\pm$0.11          & 48.12$\pm$0.13          & 47.95$\pm$0.04          & 49.57$\pm$0.15          & 44.37$\pm$0.37         \\
TRADES        & 83.91$\pm$0.33          & 54.25$\pm$0.11          & 52.21$\pm$0.09          & 55.65$\pm$0.1          & 52.22$\pm$0.05          & 48.2$\pm$0.2          \\
FAT           & \textbf{87.72$\pm$0.14} & 46.69$\pm$0.31          & 46.81$\pm$0.3          & 47.03$\pm$0.17          & 49.66$\pm$0.38          & 43.14$\pm$0.43         \\
MART          & 83.12$\pm$0.23          & 55.43$\pm$0.16          & 53.46$\pm$0.24          & \textbf{57.06$\pm$0.2} & 51.45$\pm$0.29          & 48.13$\pm$0.31         \\
GAIRAT        & 83.4$\pm$0.21           & 54.76$\pm$0.42          & 54.81$\pm$0.63          & 53.57$\pm$0.31          & 38.71$\pm$0.26          & 31.25$\pm$0.44         \\ 
PoE           & 85.41$\pm$0.29          & 55.2$\pm$0.37           & 55.07$\pm$0.24          & 54.33$\pm$0.32          & 49.25$\pm$0.16          & 46.17$\pm$0.35         \\ 
PoE (TRADES)  & 83.57$\pm$0.31          & 53.88$\pm$0.45          & 53.82$\pm$0.27          & 55.01$\pm$0.18          & 52.72$\pm$0.65          & 49.2$\pm$0.24         \\ 
CutMix (with WA) & 81.26$\pm$0.44       & 52.77$\pm$0.33          & 52.55$\pm$0.25          & 53.01$\pm$0.25          & 50.01$\pm$0.55          & 47.38$\pm$0.36         \\ \hline\hline
\textbf{SEAT} & 83.7$\pm$0.13           & \textbf{56.02$\pm$0.11} & \textbf{55.97$\pm$0.07} & \textbf{57.13$\pm$0.12} & \textbf{54.38$\pm$0.1} & \textbf{51.3$\pm$0.26} \\
\textbf{SEAT+CutMix} & 81.53$\pm$0.31   & 55.3$\pm$0.27           & 54.82$\pm$0.18          & 56.41$\pm$0.17          & 53.83$\pm$0.31          & 49.1$\pm$0.44
\end{tabular}
\end{table*}
It's worth noting that PoE is unexpectedly much weaker in defending AA attack even than its members.
We guess the poor performance of FAT and GAIRAT encumber the collective though we appropriately lower the contribution score of them. If we switch to an ensemble of the outputs of TRADES with different $\lambda$s, the performance will be slightly improved. That demonstrates that the robustness of every candidate model will affect the performance of the ensemble. The negligible computational burden estimated in Appendix A.5 also demonstrates the superiority of our on-the-fly parameter-averaged method of SEAT.

We also evaluate the SEAT method on WRN-32-10. Results are shown in Table \ref{tab:WRN}.
We notice that the superiority of SEAT appears to be enhanced when using WRN-32-10. The gap between SEAT and advanced defense methods enlarges to at least $\sim4\%$.
In the setting with data augmentation, when compared with the results of ResNet18, SEAT combined with CutMix gains much higher performance on robust accuracy against several attacks as the model becomes larger. The observation given by \cite{rebuffi2021fixing} shows that additional generated data would overload the model with low capacity. When it comes to ours, we can further infer from a marginal and even a negative effect shown in Table \ref{tab:Resnet18} that the model with low capacity cannot benefit from both external and internal generated data. This phenomenon springs from the fact that CutMix techniques tend to produce augmented views that are far away from the original image they augment, which means that they are a little hard for small neural networks to learn.
\renewcommand{\arraystretch}{0.9}
\begin{table*}[!t]
\caption{Comparison of our algorithm with different defense methods using WRN-32-10 on CIFAR-10. The maximum perturbation is $\varepsilon=8/255$. Average accuracy rates (in \%) and standard deviations have shown that SEAT also shows a great improvement on robustness.}\label{tab:WRN}
\centering
\footnotesize
\begin{tabular}{l|cccccc}
Method & NAT   & $\operatorname{PGD}^{20}$          & $\operatorname{PGD}^{100}$ & MIM   & CW    & AA    \\ \hline\hline
AT     & 87.32$\pm$0.21 & 49.01$\pm$0.33 & 48.83$\pm$0.27  & 48.25$\pm$0.17 & 52.8$\pm$0.25  & 48.17$\pm$0.48 \\
TRADES & 85.11$\pm$0.77 & 54.58$\pm$0.49 & 54.82$\pm$0.38  & 55.67$\pm$0.31 & 54.91$\pm$0.21 & 52.19$\pm$0.44 \\
FAT    & \textbf{89.65$\pm$0.04} & 48.74$\pm$0.23 & 48.69$\pm$0.18  & 48.24$\pm$0.16 & 52.11$\pm$0.71 & 46.7$\pm$0.4  \\
MART   & 84.26$\pm$0.28 & 54.11$\pm$0.58 & 54.13$\pm$0.3  & 55.2$\pm$0.22  & 53.41$\pm$0.17 & 50.2$\pm$0.36  \\
GAIRAT & 85.92$\pm$0.69 & 58.51$\pm$0.42 & 58.48$\pm$0.34  & 58.37$\pm$0.27 & 44.31$\pm$0.22 & 39.64$\pm$1.01 \\ 
PoE    & 87.1$\pm$0.25  & 55.75$\pm$0.2 & 55.47$\pm$0.19  & 56.04$\pm$0.31 & 53.66$\pm$0.18 & 49.44$\pm$0.35         \\ 
PoE (TRADES) & 86.03$\pm$0.37          & 54.26$\pm$0.47           & 54.73$\pm$0.21          & 55.01$\pm$0.22          & 55.52$\pm$0.18          & 53.2$\pm$0.4         \\ 
CutMix (with WA) & 82.79$\pm$0.44 & 58.43$\pm$1.21 & 58.2$\pm$0.83  & 58.95$\pm$0.57 & 58.32$\pm$0.43 & 54.1$\pm$0.82 \\ \hline\hline
\textbf{SEAT}   & 86.44$\pm$0.12 & 59.84$\pm$0.2 & 59.8$\pm$0.16  & \textbf{60.87$\pm$0.1} & 58.95$\pm$0.34 & 55.67$\pm$0.22 \\
\textbf{SEAT+CutMix}   & 84.81$\pm$0.18 & \textbf{60.2$\pm$0.16}  & \textbf{60.31$\pm$0.12}   & 60.53$\pm$0.21 & \textbf{59.46$\pm$0.24} & \textbf{56.03$\pm$0.36}
\end{tabular}
\end{table*}

\subsection{Ablation Study}\label{sec_ablations} 
In this section, we perform several ablation experiments to investigate how different aspects of SEAT influence its effectiveness. If not specified otherwise, the experiments are conducted on CIFAR-10 using ResNet18.
\begin{figure}[!ht]
    \begin{minipage}[t]{.32\textwidth}
        \centering
        \includegraphics[width=\textwidth]{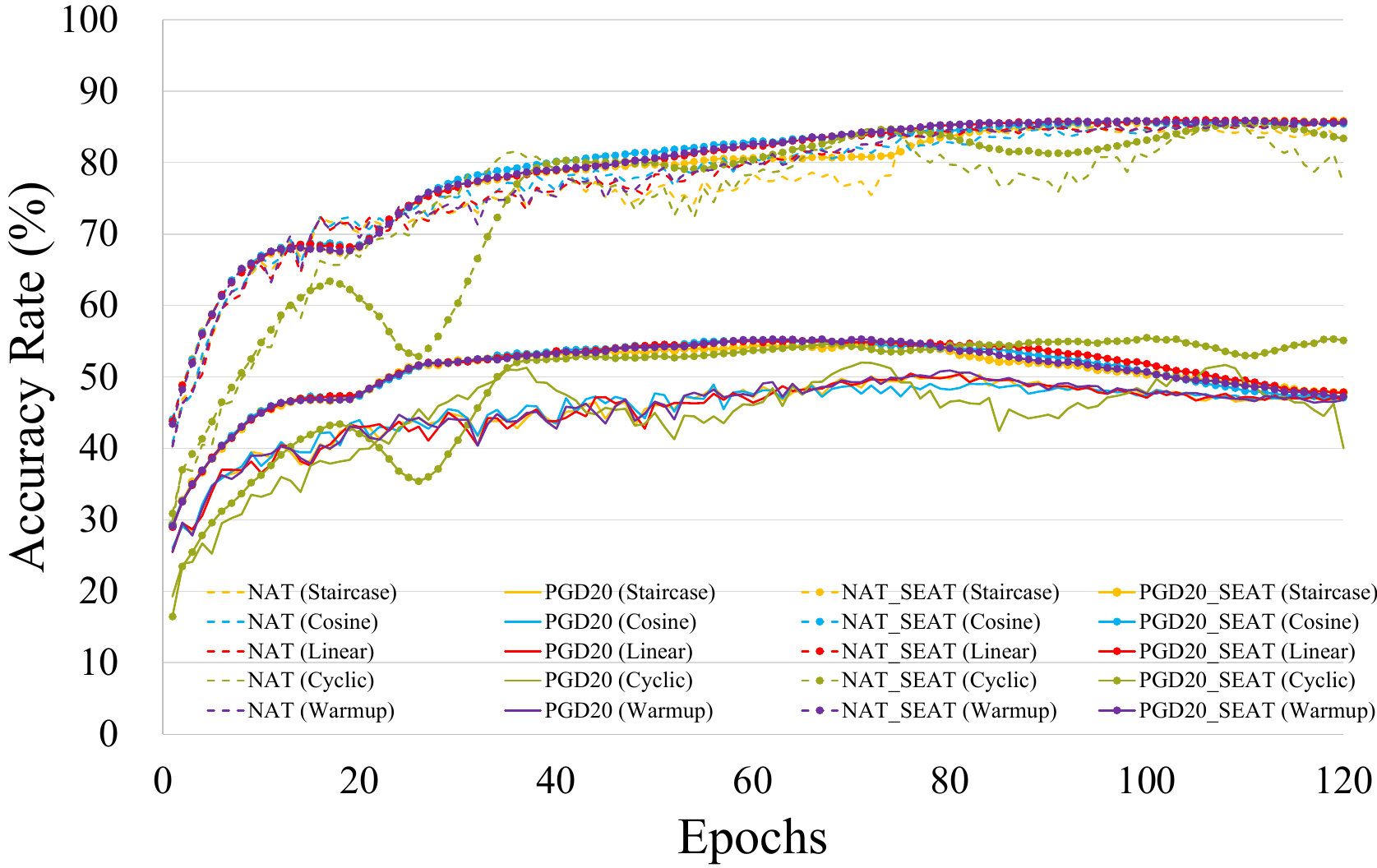}
        \subcaption{Different learning rate strategies}\label{fig:lr}
    \end{minipage}
    \hfill
    \begin{minipage}[t]{.32\textwidth}
        \centering
        \includegraphics[width=\textwidth]{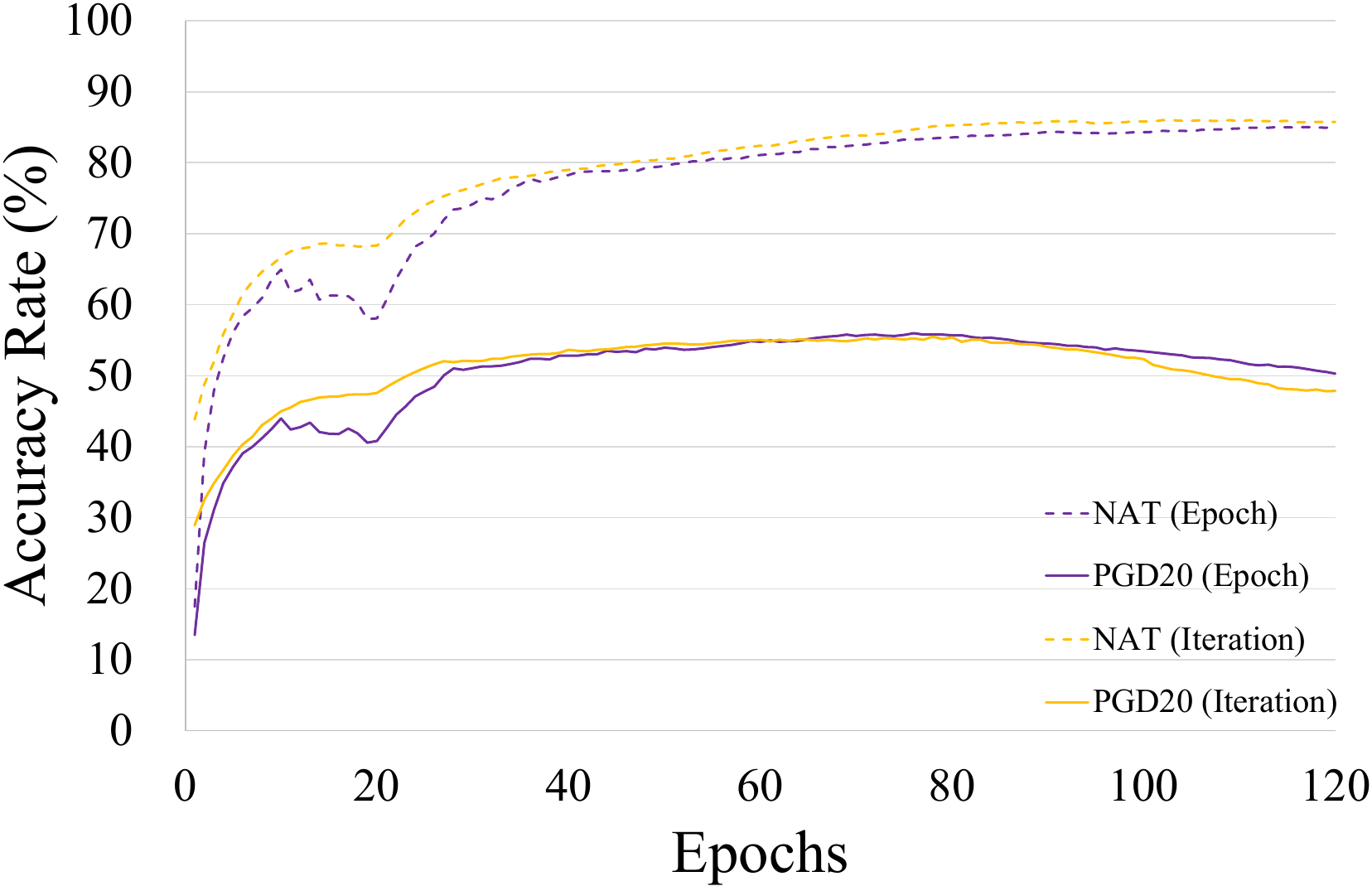}
        \subcaption{Epoch v.s. Iteration}\label{fig:epoch_iter}
    \end{minipage}
    \hfill
    \begin{minipage}[t]{.32\textwidth}
        \centering
        \includegraphics[width=\textwidth]{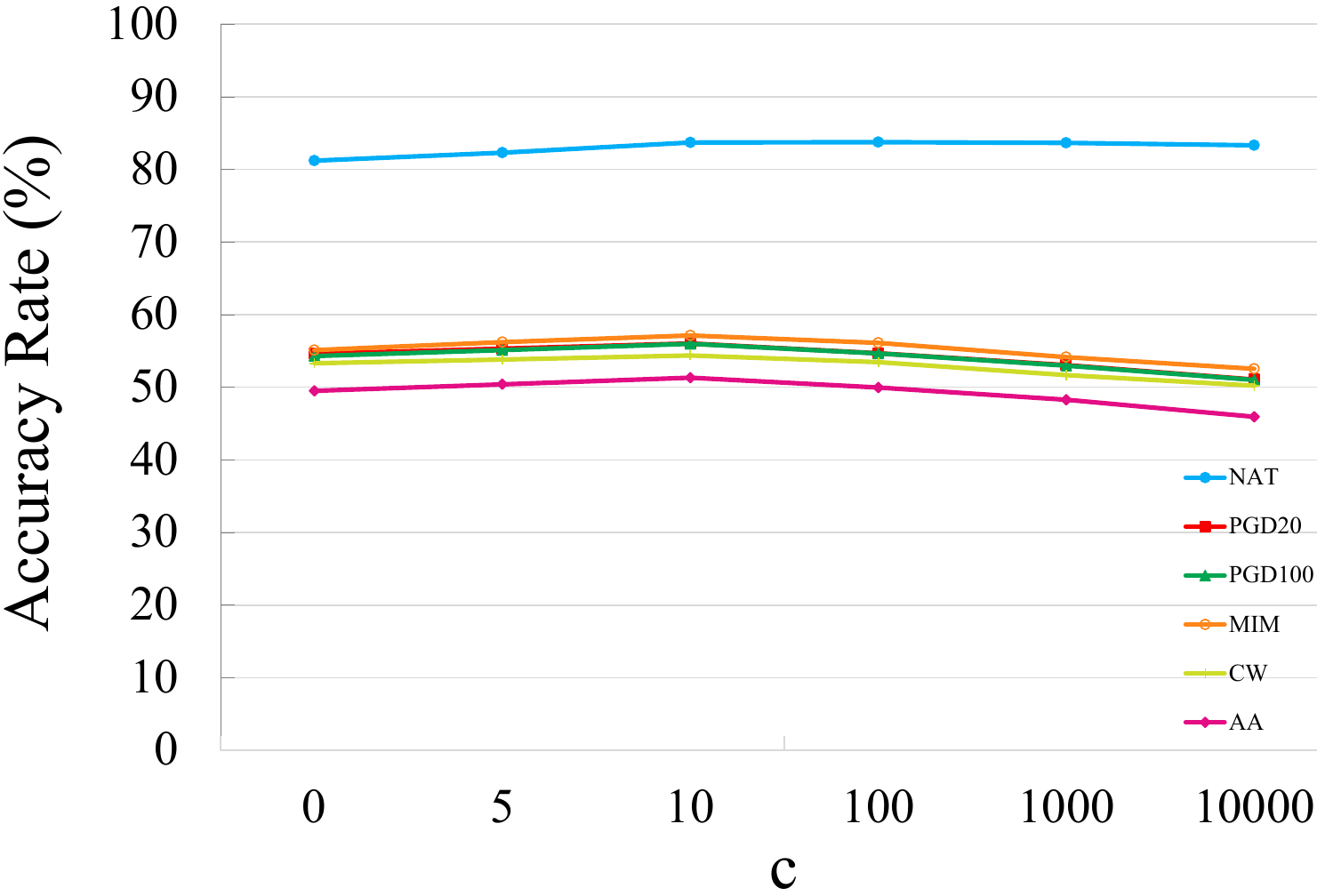}
        \subcaption{Length of initial safeguard}\label{fig:safeguard}
    \end{minipage}  
    \vspace{-0.1 in}
    \caption{The overall ablation study of SEAT. The dashed line represents the natural accuracy while the solid line represents the robust accuracy.}\label{fig:ablations}
\end{figure}

\textbf{Different Learning Rate Strategies.} 
The staircase learning rate technique is always applied to update weights in commonly used neural networks, which $\tau$ remains unchanged within the specific epochs and suddenly decreases to a certain value at some points. We also have the cosine, cyclic, and warming up learning rate schedules to compare with, in addition to the staircase and piecewise linear one deployed to SEAT. Figure \ref{fig:lr} depicts the results of this experiment using various learning rate methodologies. The natural accuracy under various learning rate schedules is reported by the top dashed lines, which all grow gradually over time. After reaching their maxima, the relative bottom solid lines fall more or less. Quantitative Results of above five strategies from best checkpoints are shown in Appendix A.3.
However, it can be observed that the red line (piecewise linear) and the blue line (cosine) decrease more slowly than the orange line (staircase) and they achieve higher robust accuracy at much later epochs than the staircase. Apparently, the piecewise linear strategy makes $\tau$ approach the threshold $\tau_{t^{\prime}}$ at a slow pace while the cosine one increases the valid candidate models that can make contributions. Likewise, the cyclic strategy gets better robust accuracy and further improves the performance in the late stage. But the effect of warming up learning rate is marginal since the homogenization of the candidate models we mentioned in Sec. 3.2 cannot be fixed by the warmup strategy. These facts bear out Proposition \ref{pro:2} from the side.

\textbf{Epoch-based v.s. Iteration-based Ensembling.} 
Next, we investigate whether the frequency of ensembling has a great effect on the robustness of the collective. Specifically, we select two scenarios to compare with. One is an epoch-based ensemble that only gathers the updated models after traversing the whole training dataset. The other is an iteration-based ensemble collecting models whenever SGD functions. Comparison between the epoch-based and the iteration-based ensembling is shown in Figure \ref{fig:epoch_iter}. It can be seen that the curves of both two schemes are nearly the same in their prime except that the iteration-based method has a slightly greater natural accuracy and degrades more slowly. 

\textbf{Influence of the Initial Safeguard.} 
We also study the effect of the length of the initial safeguard. We plot all results against different kinds of attacks with $c$ ranging from 0 to 10000 (i.e. Epoch 32). The curves in Figure \ref{fig:safeguard} show that the model ensembled with large $c$, which ignores more history models, results in weaker memorization and poor robustness. 
Actually, from the perspective of Proposition \ref{pro:2}, $t^{\prime}$ is fixed if the scheme of learning rate updating and the total training epochs are unchanged. So using a large $c$ is equivalent to squeeze the number of benign candidates before $t^{\prime}$. Similarly, a small $c$ can manually filter the early models without being well trained. However, such a drop is minor because EMA has a built-in mechanism for adjusting. 

\textbf{Mixed Strategies with SEAT for Improved Robustness.}
Finally, we report the results under different mixed strategies on CIFAR-10 dataset with ResNet18 in Table \ref{tab:combination}. Overall, SEAT+TRADES and SEAT+TRADES+CutMix do not perform well. We believe the reason is that TRADES cannot well adapt to learning strategies other than the staircase one for its history snapshots of TRADES trained under other strategies are much inferior than the ones trained under the staircase learning rate strategy. However, MART is in harmony with other strategies. The combination of SEAT+MART+CutMix achieves the best or the second-best robustness against almost all types of attacks.
\begin{table*}[!htb]
\centering
\tiny
\caption{Average robust accuracy (\%) and standard deviation under different mixed strategies on CIFAR-10 dataset with ResNet18.}\label{tab:combination}
\begin{tabular}{l|ccccccccccc}
Method                                                        & NAT                                                           & PGD20                                                         & PGD100                                                         & MIM                                                            & CW                                                            & APGDCE                                                         & APGDDLR                                                        & APGDT                                                         & FABT                                                           & Square                                                         & AA                                                            \\ \hline\hline
SEAT                                                          & \textbf{\begin{tabular}[c]{@{}c@{}}83.7\\ ±0.13\end{tabular}} & \begin{tabular}[c]{@{}c@{}}56.02\\ ±0.11\end{tabular}         & \begin{tabular}[c]{@{}c@{}}55.97\\ ±0.07\end{tabular}          & \begin{tabular}[c]{@{}c@{}}57.13\\ ±0.12\end{tabular}          & \textbf{\begin{tabular}[c]{@{}c@{}}54.38\\ ±0.1\end{tabular}} & \begin{tabular}[c]{@{}c@{}}53.87\\ ±0.17\end{tabular}          & \begin{tabular}[c]{@{}c@{}}53.35\\ ±0.24\end{tabular}          & \begin{tabular}[c]{@{}c@{}}50.88\\ ±0.27\end{tabular}         & \begin{tabular}[c]{@{}c@{}}51.41\\ ±0.37\end{tabular}          & \begin{tabular}[c]{@{}c@{}}57.77\\ ±0.22\end{tabular}          & \begin{tabular}[c]{@{}c@{}}51.3\\ ±0.26\end{tabular}          \\
\begin{tabular}[c]{@{}l@{}}SEAT\\ +TRADES\end{tabular}        & \begin{tabular}[c]{@{}c@{}}81.21\\ ±0.44\end{tabular}         & \begin{tabular}[c]{@{}c@{}}57.05\\ ±0.28\end{tabular}         & \begin{tabular}[c]{@{}c@{}}57.0\\ ±0.15\end{tabular}           & \textbf{\begin{tabular}[c]{@{}c@{}}57.92\\ ±0.12\end{tabular}} & \begin{tabular}[c]{@{}c@{}}52.75\\ ±0.09\end{tabular}         & \begin{tabular}[c]{@{}c@{}}50.75\\ ±0.25\end{tabular}          & \begin{tabular}[c]{@{}c@{}}50.36\\ ±0.29\end{tabular}          & \begin{tabular}[c]{@{}c@{}}48.56\\ ±0.37\end{tabular}         & \begin{tabular}[c]{@{}c@{}}49.45\\ ±0.65\end{tabular}          & \begin{tabular}[c]{@{}c@{}}54.45\\ ±0.58\end{tabular}          & \begin{tabular}[c]{@{}c@{}}49.91\\ ±0.17\end{tabular}         \\
\begin{tabular}[c]{@{}l@{}}SEAT\\ +MART\end{tabular}          & \begin{tabular}[c]{@{}c@{}}78.94\\ ±0.14\end{tabular}         & \begin{tabular}[c]{@{}c@{}}56.92\\ ±0.46\end{tabular}         & \begin{tabular}[c]{@{}c@{}}56.88\\ ±0.29\end{tabular}          & \begin{tabular}[c]{@{}c@{}}57.24\\ ±0.48\end{tabular}          & \begin{tabular}[c]{@{}c@{}}52.09\\ ±0.66\end{tabular}         & \begin{tabular}[c]{@{}c@{}}54.06\\ ±0.56\end{tabular}          & \begin{tabular}[c]{@{}c@{}}53.77\\ ±0.25\end{tabular}          & \begin{tabular}[c]{@{}c@{}}51.03\\ ±0.44\end{tabular}         & \begin{tabular}[c]{@{}c@{}}51.11\\ ±0.41\end{tabular}          & \begin{tabular}[c]{@{}c@{}}57.76\\ ±0.08\end{tabular}          & \begin{tabular}[c]{@{}c@{}}51.7\\ ±0.35\end{tabular}          \\
\begin{tabular}[c]{@{}l@{}}SEAT\\ +TRADES+CutMix\end{tabular} & \begin{tabular}[c]{@{}c@{}}78.22\\ ±0.33\end{tabular}         & \begin{tabular}[c]{@{}c@{}}57.14\\ ±0.21\end{tabular}         & \begin{tabular}[c]{@{}c@{}}57.09\\ ±0.45\end{tabular}          & \begin{tabular}[c]{@{}c@{}}57.11\\ ±0.39\end{tabular}          & \begin{tabular}[c]{@{}c@{}}53.17\\ ±0.41\end{tabular}         & \begin{tabular}[c]{@{}c@{}}52.67\\ ±0.82\end{tabular}          & \begin{tabular}[c]{@{}c@{}}51.86\\ ±0.5\end{tabular}           & \begin{tabular}[c]{@{}c@{}}50.91\\ ±0.17\end{tabular}         & \begin{tabular}[c]{@{}c@{}}50.23\\ ±0.29\end{tabular}          & \begin{tabular}[c]{@{}c@{}}55.12\\ ±0.46\end{tabular}          & \begin{tabular}[c]{@{}c@{}}50.01\\ ±0.44\end{tabular}         \\
\begin{tabular}[c]{@{}l@{}}SEAT\\ +MART+CutMix\end{tabular}   & \begin{tabular}[c]{@{}c@{}}75.87\\ ±0.24\end{tabular}         & \textbf{\begin{tabular}[c]{@{}c@{}}57.3\\ ±0.16\end{tabular}} & \textbf{\begin{tabular}[c]{@{}c@{}}57.29\\ ±0.43\end{tabular}} & \begin{tabular}[c]{@{}c@{}}57.47\\ ±0.18\end{tabular}          & \begin{tabular}[c]{@{}c@{}}53.13\\ ±0.2\end{tabular}          & \textbf{\begin{tabular}[c]{@{}c@{}}54.33\\ ±0.33\end{tabular}} & \textbf{\begin{tabular}[c]{@{}c@{}}53.98\\ ±0.61\end{tabular}} & \textbf{\begin{tabular}[c]{@{}c@{}}51.2\\ ±0.73\end{tabular}} & \textbf{\begin{tabular}[c]{@{}c@{}}51.42\\ ±0.17\end{tabular}} & \textbf{\begin{tabular}[c]{@{}c@{}}58.01\\ ±0.08\end{tabular}} & \textbf{\begin{tabular}[c]{@{}c@{}}52.1\\ ±0.22\end{tabular}}
\end{tabular}
\end{table*}

\section{Conclusion}

In this paper, we propose a simple but powerful method called Self-Ensemble Adversarial Training (SEAT), which unites states of every history model on the optimization trajectory through the process of adversarial training. Compared with the standard ensemble method, SEAT only needs training once and has a better reusability. 
Besides, we give a theoretical explanation of the difference between the above two ensemble methods and visualize the change of loss landscape caused by SEAT. Furthermore, we analyze a subtle but fatal intrinsic issue in the learning rate strategy for the self-ensemble model, which causes the deterioration of the weight-ensembled method. 
Extensive experiments validate the effectiveness of the proposed SEAT method.

\section*{Acknowledgement}
Yisen Wang is partially supported by the National Natural Science Foundation of China under Grant
62006153, Project 2020BD006 supported by PKU-Baidu Fund, and Open Research Projects of Zhejiang Lab (No. 2022RC0AB05).

\bibliography{iclr2022_conference}
\bibliographystyle{iclr2022_conference}

\newpage

\appendix

\section{Further Experiments}
Here we adopt ResNet18 and / or WRN-32-10 as the backbone model with the same experimental setup as in Sec. 4.1, where we reported the natural accuracy (NAT), PGD-20 and PGD-100 attack (PGD), MIM (PGD with a momentum term), CW attack and each component of AutoAttack. All the experiments are conducted for 5 individual trials and we also report their standard deviations. All the methods were realized by Pytorch 1.5, where we used a single NVIDIA GeForce RTX 3090 GPU. 

\subsection{Robustness against Components of AutoAttack} 
To broadly demonstrate the robustness of our proposal, we conducted experiments against each component of AutoAttack. We perform each component of AA on CIFAR-10 dataset with both ResNet18 and WRN-32-10, including three parameter-free versions of PGD with the CE, DLR, targeted-CE loss with 9 target classes loss ($\operatorname{APGD}_{CE}$, $\operatorname{APGD}_{DLR}$, $\operatorname{APGD}_{T}$), the targeted version of FAB ($\operatorname{FAB}_{T}$) and an existing complementary Square \citep{ACFH2020square}. Results are shown in the following Table \ref{tab:AA_components}. And it is obvious that our SEAT outperforms other methods against all components of AA. 

\begin{table*}[!htb]
\centering
\tiny
\caption{Average robust accuracy (\%) and standard deviation against each component of AA on CIFAR-10 dataset with ResNet18 and WRN-32-10.}\label{tab:AA_components}
\begin{tabular}{l|ccccc|ccccc}
       & \multicolumn{5}{c|}{ResNet18}                                                                                                   & \multicolumn{5}{c}{WRN-32-10}                                                                                                     \\
       & $\operatorname{APGD}_{CE}$                                                         & $\operatorname{APGD}_{DLR}$                                                        & $\operatorname{APGD}_{T}$                                                          & $\operatorname{FAB}_{T}$                                                           & Square                                         & $\operatorname{APGD}_{CE}$                                                         & $\operatorname{APGD}_{DLR}$                                                        & $\operatorname{APGD}_{T}$                                                          & $\operatorname{FAB}_{T}$                                                           & Square                                     \\ \hline\hline
AT     & \begin{tabular}[c]{@{}c@{}}47.47\\ $\pm$0.35\end{tabular}          & \begin{tabular}[c]{@{}c@{}}48.57\\ $\pm$0.18\end{tabular}          & \begin{tabular}[c]{@{}c@{}}45.14\\ $\pm$0.31\end{tabular}          & \begin{tabular}[c]{@{}c@{}}46.17\\ $\pm$0.11\end{tabular}          & \begin{tabular}[c]{@{}c@{}}54.21\\ $\pm$0.15\end{tabular}          & \begin{tabular}[c]{@{}c@{}}49.17\\ $\pm$0.26\end{tabular}          & \begin{tabular}[c]{@{}c@{}}50.09\\ $\pm$0.36\end{tabular}          & \begin{tabular}[c]{@{}c@{}}47.34\\ $\pm$0.33\end{tabular}          & \begin{tabular}[c]{@{}c@{}}48.00\\ $\pm$0.43\end{tabular}          & \begin{tabular}[c]{@{}c@{}}56.5\\ $\pm$0.18\end{tabular}           \\
TRADES & \begin{tabular}[c]{@{}c@{}}53.47\\ $\pm$0.21\end{tabular}          & \begin{tabular}[c]{@{}c@{}}50.89\\ $\pm$0.26\end{tabular}          & \begin{tabular}[c]{@{}c@{}}47.93\\ $\pm$0.36\end{tabular}          & \begin{tabular}[c]{@{}c@{}}48.53\\ $\pm$0.43\end{tabular}          & \begin{tabular}[c]{@{}c@{}}55.75\\ $\pm$0.21\end{tabular}          & \begin{tabular}[c]{@{}c@{}}55.38\\ $\pm$0.43\end{tabular}          & \begin{tabular}[c]{@{}c@{}}55.55\\ $\pm$0.42\end{tabular}          & \begin{tabular}[c]{@{}c@{}}52.2\\ $\pm$0.13\end{tabular}           & \begin{tabular}[c]{@{}c@{}}53.11\\ $\pm$0.72\end{tabular}          & \begin{tabular}[c]{@{}c@{}}59.47\\ $\pm$0.17\end{tabular}          \\
MART   & \begin{tabular}[c]{@{}c@{}}52.98\\ $\pm$0.13\end{tabular}          & \begin{tabular}[c]{@{}c@{}}50.36\\ $\pm$0.3\end{tabular}           & \begin{tabular}[c]{@{}c@{}}48.17\\ $\pm$0.72\end{tabular}          & \begin{tabular}[c]{@{}c@{}}49.39\\ $\pm$0.28\end{tabular}          & \begin{tabular}[c]{@{}c@{}}55.73\\ $\pm$0.51\end{tabular}          & \begin{tabular}[c]{@{}c@{}}55.2\\ $\pm$0.32\end{tabular}           & \begin{tabular}[c]{@{}c@{}}55.41\\ $\pm$0.4\end{tabular}           & \begin{tabular}[c]{@{}c@{}}51.99\\ $\pm$0.3\end{tabular}           & \begin{tabular}[c]{@{}c@{}}52.88\\ $\pm$0.63\end{tabular}          & \begin{tabular}[c]{@{}c@{}}59.01\\ $\pm$0.38\end{tabular}          \\ \hline\hline
SEAT   & \textbf{\begin{tabular}[c]{@{}c@{}}53.87\\ $\pm$0.17\end{tabular}} & \textbf{\begin{tabular}[c]{@{}c@{}}53.35\\ $\pm$0.24\end{tabular}} & \textbf{\begin{tabular}[c]{@{}c@{}}50.88\\ $\pm$0.27\end{tabular}} & \textbf{\begin{tabular}[c]{@{}c@{}}51.41\\ $\pm$0.37\end{tabular}} & \textbf{\begin{tabular}[c]{@{}c@{}}57.77\\ $\pm$0.22\end{tabular}} & \textbf{\begin{tabular}[c]{@{}c@{}}57.57\\ $\pm$0.18\end{tabular}} & \textbf{\begin{tabular}[c]{@{}c@{}}57.74\\ $\pm$0.29\end{tabular}} & \textbf{\begin{tabular}[c]{@{}c@{}}55.06\\ $\pm$0.27\end{tabular}} & \textbf{\begin{tabular}[c]{@{}c@{}}55.53\\ $\pm$0.36\end{tabular}} & \textbf{\begin{tabular}[c]{@{}c@{}}62.26\\ $\pm$0.23\end{tabular}}
\end{tabular}
\end{table*}

\subsection{Performance on CIFAR-100}
To further demonstrate the robustness of our proposal against adversarial attacks, we benchmark the state-of-the-art robustness with ResNet18 on CIFAR-100. We widely investigate the performance of SEAT against the PGD methods ($\operatorname{PGD}^{20}$ and $\operatorname{PGD}^{100}$), MIM, CW, AA and its all components. Results shown in Table \ref{tab:cifar100} demonstrate the effectiveness of SEAT for building a robust classifier.
\begin{table*}[!htb]
\centering
\tiny
\caption{Comparison of our algorithm with different defense methods using ResNet18 on CIFAR10. The maximum perturbation is $\varepsilon=8/255$. Average accuracy rates (in \%) and standard deviations have shown that the proposed SEAT method greatly improves the robustness of the model.}\label{tab:cifar100}
\begin{tabular}{l|ccccccccccc}
Method        & NAT                                                           & $\operatorname{PGD}^{20}$                                                          & $\operatorname{PGD}^{100}$                                                         & MIM                                                            & CW                                                             & $\operatorname{APGD}_{CE}$                                                         & $\operatorname{APGD}_{DLR}$                                                        & $\operatorname{APGD}_{T}$                                                          & $\operatorname{FAB}_{T}$                                                           & Square                                                         & AA                                                             \\ \hline\hline
AT            & \textbf{\begin{tabular}[c]{@{}c@{}}60.1\\ $\pm$0.35\end{tabular}} & \begin{tabular}[c]{@{}c@{}}28.22\\ $\pm$0.3\end{tabular}           & \begin{tabular}[c]{@{}c@{}}28.27\\ $\pm$0.12\end{tabular}          & \begin{tabular}[c]{@{}c@{}}28.31\\ $\pm$0.41\end{tabular}          & \begin{tabular}[c]{@{}c@{}}24.87\\ $\pm$0.51\end{tabular}          & \begin{tabular}[c]{@{}c@{}}26.63\\ $\pm$0.29\end{tabular}          & \begin{tabular}[c]{@{}c@{}}24.13\\ $\pm$0.22\end{tabular}          & \begin{tabular}[c]{@{}c@{}}21.98\\ $\pm$0.3\end{tabular}           & \begin{tabular}[c]{@{}c@{}}23.87\\ $\pm$0.21\end{tabular}          & \begin{tabular}[c]{@{}c@{}}27.93\\ $\pm$0.12\end{tabular}          & \begin{tabular}[c]{@{}c@{}}23.91\\ $\pm$0.41\end{tabular}          \\
TRADES        & \begin{tabular}[c]{@{}c@{}}59.93\\ $\pm$0.46\end{tabular}         & \begin{tabular}[c]{@{}c@{}}29.9\\ $\pm$0.41\end{tabular}           & \begin{tabular}[c]{@{}c@{}}29.88\\ $\pm$0.11\end{tabular}          & \begin{tabular}[c]{@{}c@{}}29.55\\ $\pm$0.25\end{tabular}          & \begin{tabular}[c]{@{}c@{}}26.14\\ $\pm$0.21\end{tabular}          & \begin{tabular}[c]{@{}c@{}}27.93\\ $\pm$0.44\end{tabular}          & \begin{tabular}[c]{@{}c@{}}25.43\\ $\pm$0.29\end{tabular}          & \begin{tabular}[c]{@{}c@{}}23.72\\ $\pm$0.45\end{tabular}          & \begin{tabular}[c]{@{}c@{}}25.16\\ $\pm$0.15\end{tabular}          & \begin{tabular}[c]{@{}c@{}}30.03\\ $\pm$0.32\end{tabular}          & \begin{tabular}[c]{@{}c@{}}24.72\\ $\pm$0.37\end{tabular}          \\
MART          & \begin{tabular}[c]{@{}c@{}}57.24\\ $\pm$0.64\end{tabular}         & \begin{tabular}[c]{@{}c@{}}30.62\\ $\pm$0.37\end{tabular}          & \begin{tabular}[c]{@{}c@{}}30.62\\ $\pm$0.17\end{tabular}          & \begin{tabular}[c]{@{}c@{}}30.83\\ $\pm$0.28\end{tabular}          & \begin{tabular}[c]{@{}c@{}}26.3\\ $\pm$0.29\end{tabular}           & \begin{tabular}[c]{@{}c@{}}29.91\\ $\pm$0.07\end{tabular}          & \begin{tabular}[c]{@{}c@{}}26.32\\ $\pm$0.24\end{tabular}          & \begin{tabular}[c]{@{}c@{}}24.28\\ $\pm$0.49\end{tabular}          & \begin{tabular}[c]{@{}c@{}}24.86\\ $\pm$0.66\end{tabular}          & \begin{tabular}[c]{@{}c@{}}28.28\\ $\pm$0.39\end{tabular}          & \begin{tabular}[c]{@{}c@{}}24.27\\ $\pm$0.21\end{tabular}          \\ \hline\hline
\textbf{SEAT} & \begin{tabular}[c]{@{}c@{}}56.28\\ $\pm$0.33\end{tabular}         & \textbf{\begin{tabular}[c]{@{}c@{}}32.15\\ $\pm$0.17\end{tabular}} & \textbf{\begin{tabular}[c]{@{}c@{}}32.12\\ $\pm$0.26\end{tabular}} & \textbf{\begin{tabular}[c]{@{}c@{}}32.62\\ $\pm$0.15\end{tabular}} & \textbf{\begin{tabular}[c]{@{}c@{}}29.68\\ $\pm$0.26\end{tabular}} & \textbf{\begin{tabular}[c]{@{}c@{}}30.97\\ $\pm$0.18\end{tabular}} & \textbf{\begin{tabular}[c]{@{}c@{}}29.62\\ $\pm$0.22\end{tabular}} & \textbf{\begin{tabular}[c]{@{}c@{}}26.88\\ $\pm$0.23\end{tabular}} & \textbf{\begin{tabular}[c]{@{}c@{}}27.71\\ $\pm$0.24\end{tabular}} & \textbf{\begin{tabular}[c]{@{}c@{}}32.35\\ $\pm$0.34\end{tabular}} & \textbf{\begin{tabular}[c]{@{}c@{}}27.87\\ $\pm$0.24\end{tabular}}
\end{tabular}
\end{table*}

\subsection{Different Learning Rate Strategies}
Apart from showing the curve of different learning rate schedule in Figure 3 (a) in Sec. 4.3, we also report final results in Table \ref{tab:lr}. The effect of warming up learning rate is marginal. When compared with the staircase one, the warmup strategy cannot generate diverse models in the later stages so the homogenization of the candidate models we mentioned in Sec. 3.2 cannot be fixed by the warmup strategy. On the contrary, those methods like cosine / linear / cyclic that provide relatively diverse models in the later stages can mitigate the issue, accounting for more robust ensemble models.
\begin{table*}[!htb]
\centering
\tiny
\caption{Average robust accuracy (\%) under different learning strategies on CIFAR-10 dataset with ResNet18.}\label{tab:lr}
\begin{tabular}{l|ccccccccccc}
Method           & NAT                                                           & $\operatorname{PGD}^{20}$                                                          & $\operatorname{PGD}^{100}$                                                         & MIM                                                            & CW                                                             & $\operatorname{APGD}_{CE}$                                                         & $\operatorname{APGD}_{DLR}$                                                        & $\operatorname{APGD}_{T}$                                                          & $\operatorname{FAB}_{T}$                                                           & Square                                                         & AA    \\\hline\hline
SEAT (Staircase) & 80.91 & 54.58 & 54.56  & 54.47 & 49.71 & 52.39  & 48.01   & 45.83 & 45.11 & 53.64  & 45.85 \\
SEAT (Cosine)    & 83.0  & 55.09 & 55.16  & 56.39 & 53.43 & 52.34  & 52.1    & 49.51 & 50.15 & 56.48  & 50.48 \\
SEAT (Linear)    & \textbf{83.7}  & \textbf{56.02} & \textbf{55.97}  & \textbf{57.13} & \textbf{54.38} & \textbf{53.87}  & \textbf{53.35}   & \textbf{50.88} & \textbf{51.41} & \textbf{57.77}  & \textbf{51.3}  \\
SEAT (Warmup)    & 82.74 & 55.31 & 55.35  & 56.39 & 53.26 & 53.55  & 48.94   & 45.89 & 46.6  & 54.94  & 45.82 \\
SEAT (Cyclic)    & 83.14 & \textbf{56.03} & \textbf{55.79}  & \textbf{56.99} & \textbf{54.01} & \textbf{53.72}  & \textbf{53.1}    & \textbf{50.66} & \textbf{51.02} & \textbf{57.75}  & \textbf{51.44}
\end{tabular}
\end{table*}

\subsection{Deterioration of vanilla EMA}
As shown in Sec. 3.3, the deteriorated SEAT underperforms SEAT a lot from the perspective of optimization. We also report quantitative results on both ResNet18 and WRN-32-10 shown in the Tables \ref{tab:resnet18_deterioration} and \ref{tab:WRN_deterioration}. The deteriorated one does not bring too much boost when compared to the vanilla adversarial training (except for PGD methods). A plausible explanation for the exception of PGD is that the SEAT technique produces an ensemble of individuals that are adversarially trained by PGD with the cross-entropy loss, which means that they are intrinsically good at defending the PGD attack with the cross-entropy loss and its variants even though they suffer from the deterioration. Considering results have greatly improved after using the piecewise linear learning rate strategy, it is fair to say that adjusting learning rate is effective. As we claimed in Proposition 2 and its proof, the staircase will inevitably make the self-ensemble model worsen since $\sum_{t=1}^{T} (\beta_{t}\tilde{\boldsymbol\xi}^\top)$ will gradually approach to zero, meaning the difference between $\bar{f}_{\mathcal{F}}(x,y)$ and $f_{\tilde{\boldsymbol\theta}}(x, y)$ achieves the second order of smallness. 
\begin{table*}[!htb]
\centering
\tiny
\caption{Average robust accuracy (\%) and standard deviation on CIFAR-10 dataset with ResNet18.}\label{tab:resnet18_deterioration}
\begin{tabular}{l|ccccccccccc}
Method              & NAT                                                           & $\operatorname{PGD}^{20}$                                                          & $\operatorname{PGD}^{100}$                                                         & MIM                                                            & CW                                                             & $\operatorname{APGD}_{CE}$                                                         & $\operatorname{APGD}_{DLR}$                                                        & $\operatorname{APGD}_{T}$                                                          & $\operatorname{FAB}_{T}$                                                           & Square                                                         & AA                                                    \\ \hline\hline
AT                  & \textbf{\begin{tabular}[c]{@{}c@{}}84.32\\ $\pm$0.23\end{tabular}} & \begin{tabular}[c]{@{}c@{}}48.29\\ $\pm$0.11\end{tabular} & \begin{tabular}[c]{@{}c@{}}48.12\\ $\pm$0.13\end{tabular} & \begin{tabular}[c]{@{}c@{}}47.95\\ $\pm$0.04\end{tabular} & \begin{tabular}[c]{@{}c@{}}49.57\\ $\pm$0.15\end{tabular} & \begin{tabular}[c]{@{}c@{}}47.47\\ $\pm$0.35\end{tabular} & \begin{tabular}[c]{@{}c@{}}48.57\\ $\pm$0.18\end{tabular} & \begin{tabular}[c]{@{}c@{}}45.14\\ $\pm$0.31\end{tabular} & \begin{tabular}[c]{@{}c@{}}46.17\\ $\pm$0.11\end{tabular} & \begin{tabular}[c]{@{}c@{}}54.21\\ $\pm$0.25\end{tabular} & \begin{tabular}[c]{@{}c@{}}44.37\\ $\pm$0.37\end{tabular} \\\hline\hline
\begin{tabular}[c]{@{}l@{}}SEAT\\ (deteriorated)\end{tabular} & \begin{tabular}[c]{@{}c@{}}80.91\\ $\pm$0.38\end{tabular} & \begin{tabular}[c]{@{}c@{}}54.58\\ $\pm$0.71\end{tabular} & \begin{tabular}[c]{@{}c@{}}54.56\\ $\pm$0.29\end{tabular} & \begin{tabular}[c]{@{}c@{}}54.47\\ $\pm$0.39\end{tabular} & \begin{tabular}[c]{@{}c@{}}49.71\\ $\pm$0.41\end{tabular} & \begin{tabular}[c]{@{}c@{}}52.39\\ $\pm$0.26\end{tabular} & \begin{tabular}[c]{@{}c@{}}48.01\\ $\pm$0.18\end{tabular} & \begin{tabular}[c]{@{}c@{}}45.83\\ $\pm$0.52\end{tabular} & \begin{tabular}[c]{@{}c@{}}45.11\\ $\pm$0.23\end{tabular} & \begin{tabular}[c]{@{}c@{}}53.64\\ $\pm$0.44\end{tabular} & \begin{tabular}[c]{@{}c@{}}45.85\\ $\pm$0.19\end{tabular} \\
SEAT                & \begin{tabular}[c]{@{}c@{}}83.7\\ $\pm$0.13\end{tabular}  & \textbf{\begin{tabular}[c]{@{}c@{}}56.02\\ $\pm$0.11\end{tabular}} & \textbf{\begin{tabular}[c]{@{}c@{}}55.97\\ $\pm$0.07\end{tabular}} & \textbf{\begin{tabular}[c]{@{}c@{}}57.13\\ $\pm$0.12\end{tabular}} & \textbf{\begin{tabular}[c]{@{}c@{}}54.38\\ $\pm$0.1\end{tabular}}  & \textbf{\begin{tabular}[c]{@{}c@{}}53.87\\ $\pm$0.17\end{tabular}} & \textbf{\begin{tabular}[c]{@{}c@{}}53.35\\ $\pm$0.24\end{tabular}} & \textbf{\begin{tabular}[c]{@{}c@{}}50.88\\ $\pm$0.27\end{tabular}} & \textbf{\begin{tabular}[c]{@{}c@{}}51.41\\ $\pm$0.37\end{tabular}} & \textbf{\begin{tabular}[c]{@{}c@{}}57.77\\ $\pm$0.22\end{tabular}} & \textbf{\begin{tabular}[c]{@{}c@{}}51.3\\ $\pm$0.26\end{tabular} }
\end{tabular}
\end{table*}

\begin{table*}[!htb]
\centering
\tiny
\caption{Average robust accuracy (\%) and standard deviation on CIFAR-10 dataset with WRN-32-10.}\label{tab:WRN_deterioration}
\begin{tabular}{l|ccccccccccc}
Method              & NAT                                                           & $\operatorname{PGD}^{20}$                                                          & $\operatorname{PGD}^{100}$                                                         & MIM                                                            & CW                                                             & $\operatorname{APGD}_{CE}$                                                         & $\operatorname{APGD}_{DLR}$                                                        & $\operatorname{APGD}_{T}$                                                          & $\operatorname{FAB}_{T}$                                                           & Square                                                         & AA                                                             \\ \hline\hline
AT                  & \textbf{\begin{tabular}[c]{@{}c@{}}87.32\\ $\pm$0.21\end{tabular}} & \begin{tabular}[c]{@{}c@{}}49.01\\ $\pm$0.33\end{tabular}         & \begin{tabular}[c]{@{}c@{}}48.83\\ $\pm$0.27\end{tabular}         & \begin{tabular}[c]{@{}c@{}}48.25\\ $\pm$0.17\end{tabular}         & \begin{tabular}[c]{@{}c@{}}52.8\\ $\pm$0.25\end{tabular}           & \begin{tabular}[c]{@{}c@{}}54.17\\ $\pm$0.26\end{tabular}          & \begin{tabular}[c]{@{}c@{}}53.09\\ $\pm$0.36\end{tabular}          & \begin{tabular}[c]{@{}c@{}}48.34\\ $\pm$0.33\end{tabular}          & \begin{tabular}[c]{@{}c@{}}49.00\\ $\pm$0.43\end{tabular}          & \begin{tabular}[c]{@{}c@{}}57.5\\ $\pm$0.18\end{tabular}           & \begin{tabular}[c]{@{}c@{}}48.17\\ $\pm$0.48\end{tabular}          \\\hline\hline
\begin{tabular}[c]{@{}l@{}}SEAT\\ (deteriorated)\end{tabular} & \begin{tabular}[c]{@{}c@{}}85.28\\ $\pm$0.42\end{tabular}          & \begin{tabular}[c]{@{}c@{}}55.68\\ $\pm$0.42\end{tabular}         & \begin{tabular}[c]{@{}c@{}}55.57\\ $\pm$0.19\end{tabular}         & \begin{tabular}[c]{@{}c@{}}55.6\\ $\pm$0.23\end{tabular}          & \begin{tabular}[c]{@{}c@{}}53.01\\ $\pm$0.41\end{tabular}          & \begin{tabular}[c]{@{}c@{}}54.12\\ $\pm$0.54\end{tabular}          & \begin{tabular}[c]{@{}c@{}}53.54\\ $\pm$0.28\end{tabular}          & \begin{tabular}[c]{@{}c@{}}49.95\\ $\pm$0.67\end{tabular}          & \begin{tabular}[c]{@{}c@{}}50.02\\ $\pm$0.75\end{tabular}          & \begin{tabular}[c]{@{}c@{}}57.81\\ $\pm$0.33\end{tabular}          & \begin{tabular}[c]{@{}c@{}}49.96\\ $\pm$0.31\end{tabular}          \\
\textbf{SEAT}       & \begin{tabular}[c]{@{}c@{}}86.44\\ $\pm$0.12\end{tabular}          & \textbf{\begin{tabular}[c]{@{}c@{}}59.84\\ $\pm$0.2\end{tabular}} & \textbf{\begin{tabular}[c]{@{}c@{}}59.8\\ $\pm$0.16\end{tabular}} & \textbf{\begin{tabular}[c]{@{}c@{}}60.87\\ $\pm$0.1\end{tabular}} & \textbf{\begin{tabular}[c]{@{}c@{}}58.95\\ $\pm$0.34\end{tabular}} & \textbf{\begin{tabular}[c]{@{}c@{}}57.57\\ $\pm$0.18\end{tabular}} & \textbf{\begin{tabular}[c]{@{}c@{}}57.74\\ $\pm$0.29\end{tabular}} & \textbf{\begin{tabular}[c]{@{}c@{}}55.06\\ $\pm$0.27\end{tabular}} & \textbf{\begin{tabular}[c]{@{}c@{}}55.53\\ $\pm$0.36\end{tabular}} & \textbf{\begin{tabular}[c]{@{}c@{}}62.26\\ $\pm$0.23\end{tabular}} & \textbf{\begin{tabular}[c]{@{}c@{}}55.67\\ $\pm$0.22\end{tabular}}
\end{tabular}
\end{table*}

\subsection{Computational Complexity for SEAT}
To demonstrate the efficiency of the SEAT method, we use the number of Multiply-Accumulate operations (MACs) in Giga (G) to compute the theoretical amount of multiply-add operations in DNNs, roughly GMACs = 0.5 * GFLOPs. Besides, we also provide the actual running time. As shown in Table \ref{tab:MACs}, the SEAT method takes negligible MACs and training time when compared with standard adversarial training.

\begin{table*}[!htb]
\centering
\caption{Evaluation of time complexity of SEAT. Here we use the number of Multiply-Accumulate operations  (MACs) in Giga (G) to measure the running time complexity. And we also compute the actual training time with or without the SEAT method using ResNet18 and WRN-32-10 on a single NVIDIA GeForce RTX 3090 GPU.}\label{tab:MACs}
\begin{tabular}{l|cc}
Method           & MACs (G) & Training Time (mins) \\ \hline\hline
ResNet18 (AT)    & 0.56     & 272                  \\
ResNet18 (SEAT)  & 0.59     & 273                  \\
WRN-32-10 (AT)   & 6.67     & 1534                 \\
WRN-32-10 (SEAT) & 6.81     & 1544                
\end{tabular}
\end{table*}

\section{Proofs of theoretical results}
\subsection{Proof of Proposition 1}
\begin{prop}
\label{pro:1_re}
(Restated) Let $f_{\boldsymbol\theta}\left(\cdot\right)$ denote the predictions of a neural network parametrized by weights $\boldsymbol\theta$. Assuming that $\forall \boldsymbol\theta \in \Theta$, $f_{\boldsymbol\theta}\left(\cdot\right)$ is continuous and $\forall (x, y)\in \mathbb{D}$, $f_{\boldsymbol\theta}(x, y)$ is at least twice differentiable. Consider two points $\boldsymbol\theta_t, \tilde{\boldsymbol\theta} \in\Theta$ in the weight space and let $\boldsymbol\xi=\boldsymbol\theta_t-\tilde{\boldsymbol\theta}$, for $t\in \left\{1,2,\cdots,T\right\}$, the difference between $\bar{f}_{\mathcal{F}}(x,y)$ and $f_{\tilde{\boldsymbol\theta}}(x, y)$ is of the second order of smallness if and only if $\sum_{t=1}^{T} (\beta_{t}\boldsymbol\xi^\top)=\boldsymbol 0$.
\end{prop}

\begin{proof}
For the sake of the twice differentiability of $f_{\boldsymbol\theta}(x, y)$, based on the Taylor expansion, we can fit a quadratic polynomial of $f_{\tilde{\boldsymbol\theta}}(x, y)$ to approximate the value of $f_{\boldsymbol\theta_t}(x, y)$:
\begin{equation}
f_{\boldsymbol\theta_t}(x, y) = f_{\tilde{\boldsymbol\theta}}(x, y)+\boldsymbol\xi^\top \nabla_{\boldsymbol\xi}f_{\tilde{\boldsymbol\theta}}(x, y) +\frac{1}{2} \boldsymbol\xi^{\top} \nabla_{\boldsymbol\xi}^{2}f_{\tilde{\boldsymbol\theta}}(x, y)\boldsymbol\xi+O\left(\Delta^{n}\right),
\end{equation}
where $O\left(\Delta^{n}\right)$ represents the higher-order remainder term. 
Note that the subscript $\boldsymbol\xi$ here stands for a neighborhood where the Taylor expansion approximates a function by polynomials of any point (i.e. $\tilde{\boldsymbol\theta}$) in terms of its value and derivatives. 
So the difference between the averaged prediction of candidate classifiers and the prediction of the ensembled weight classifier can be formulated as:
\begin{equation}
\begin{aligned}
\label{eqn:Taylor}
\bar{f}_{\mathcal{F}}(x,y) - f_{\tilde{\boldsymbol\theta}}(x, y) &=\sum_{t=1}^{T} \beta_{t} f_{\boldsymbol\theta_t}(x,y) - f_{\tilde{\boldsymbol\theta}}(x, y) \\
&=\bcancel{\sum_{t=1}^{T} \beta_{t}f_{\tilde{\boldsymbol\theta}}(x, y)} + \sum_{t=1}^{T} \beta_{t}\boldsymbol\xi^\top \nabla_{\boldsymbol\xi}f_{\tilde{\boldsymbol\theta}}(x, y) + \sum_{t=1}^{T} \beta_{t}O\left(\Delta^{2}\right) \bcancel{-f_{\tilde{\boldsymbol\theta}}(x, y)} \\
&=\sum_{t=1}^{T} (\beta_{t}\boldsymbol\xi^\top) \nabla_{\boldsymbol\xi}f_{\tilde{\boldsymbol\theta}}(x, y) + O(\Delta^{2}).
\end{aligned}
\end{equation}
Therefore, we can claim that the difference between $f_{\boldsymbol\theta_t}(x, y)$ and $f_{\tilde{\boldsymbol\theta}}(x, y)$ is "almost" at least of the first order of smallness except for some special cases. And we will immediately declare under which condition this difference can achieve the second order of smallness in the following proof of Theorem 1.
\end{proof}

\subsection{Proof of Theorem 1}
\begin{theorem}
\label{thm:1_re}
(Restated) Assuming that for $i,j\in \left\{1,\cdots,T\right\}$, $\boldsymbol\theta_i=\boldsymbol\theta_j$ if and only if $i=j$. The difference between the averaged prediction of multiple networks and the prediction of SEAT is of the second order of smallness if and only if $\beta_{i}=(1-\alpha)^{1-\delta(i-1)} \alpha^{T-i}$ for $i\in\left\{1,2,\cdots,T\right\}$.
\end{theorem}

\begin{proof}
According to Eqn \ref{eqn:Taylor}, we know that the second order of smallness will achieve when $\sum_{i=1}^{T} (\beta_{i}\boldsymbol\xi^\top) = \boldsymbol 0$. Thus, we continue deducing from Eqn \ref{eqn:Taylor} as:
\begin{equation}
\begin{aligned}
\sum_{i=1}^{T} (\beta_{i}\boldsymbol\xi^\top) &= \boldsymbol 0 \\
\sum_{i=1}^{T} \beta_{i}(\boldsymbol\theta_i-\tilde{\boldsymbol\theta}) &= \boldsymbol 0 \\
\sum_{i=1}^{T} \beta_{i}\boldsymbol\theta_i &= \tilde{\boldsymbol\theta} \\
\sum_{i=1}^{T} \beta_{i}\boldsymbol\theta_i &= \sum_{i=1}^{T}(1-\alpha)^{1-\delta(i-1)}\alpha^{T-i}\boldsymbol\theta_i.
\end{aligned}
\end{equation}

To get a further conclusion, we next use Mathematical Induction (MI) to prove only when $\beta_{i}=(1-\alpha)^{1-\delta(i-1)} \alpha^{T-i}$ for $i\in\left\{1,2,\cdots,T\right\}$ will lead to $\sum_{i=1}^{T} \beta_{i}\boldsymbol\theta_i = \sum_{i=1}^{T}(1-\alpha)^{1-\delta(i-1)}\alpha^{T-i}\boldsymbol\theta_i$.

\textbf{Base case:} Let $i=1$, it is clearly true that $\beta_{1}=\alpha^{T-1}$ if and only if $\beta_{1}\boldsymbol\theta_1 = \alpha^{T-1}\boldsymbol\theta_1$, hence the base case holds. 

\textbf{Inductive step:} Assume the induction hypothesis that for a particular $k$, the single case $T=k$ holds, meaning the sequence of $(\beta_{1}, \beta_{2}, \cdots, \beta_{k})$ is equal to the sequence of $((1-\alpha)^{1-\delta(0)}\alpha^{T-1}, (1-\alpha)^{1-\delta(1)}\alpha^{T-2}, \cdots, (1-\alpha)^{1-\delta(k-1)}\alpha^{T-k})$ if $\sum_{i=1}^{k} \beta_{i}\boldsymbol\theta_i = \sum_{i=1}^{k}(1-\alpha)^{1-\delta(i-1)}\alpha^{T-i}\boldsymbol\theta_i$.

For $T=k+1$, it follows that: 
\begin{equation}
\begin{aligned}
\sum_{i=1}^{k+1} \beta_{i}\boldsymbol\theta_i &= \sum_{i=1}^{k+1}(1-\alpha)^{1-\delta(i-1)}\alpha^{T-i}\boldsymbol\theta_i \\
\bcancel{\sum_{i=1}^{k} \beta_i\boldsymbol\theta_i} + \beta_{k+1}\boldsymbol\theta_{k+1} &= \bcancel{\sum_{i=1}^{k}(1-\alpha)^{1-\delta(i-1)}\alpha^{T-i}\boldsymbol\theta_i} + (1-\alpha)^{1-\delta((k+1)-1)}\alpha^{T-(k+1)}\boldsymbol\theta_{k+1} \\
\beta_{k+1}\boldsymbol\theta_{k+1} &= (1-\alpha)^{1-\delta((k+1)-1)}\alpha^{T-(k+1)}\boldsymbol\theta_{k+1} \\
\beta_{k+1} &= (1-\alpha)^{1-\delta((k+1)-1)}\alpha^{T-(k+1)}.
\end{aligned}
\end{equation}
The sequence of normalized scores at the $(k+1)$-th ensembling at left hand is $(\beta_{1}, \beta_{2}, \cdots, \beta_{k}, \beta_{k+1})$ after adding the new term $\beta_{k+1}$. Likewise, the sequence of the right hand is $(\beta_{1}, \beta_{2}, \cdots, \beta_{k})$ is equal to the sequence of $((1-\alpha)^{1-\delta(0)}\alpha^{T-1}, (1-\alpha)^{1-\delta(1)}\alpha^{T-2}, \cdots, (1-\alpha)^{1-\delta(k-1)}\alpha^{T-k})$. Because every $f_{\theta_t} \in \mathcal{F}$ is different from others and the sequence is ordered, we have $(\beta_{1}, \beta_{2}, \cdots, \beta_{k}, \beta_{k+1})=((1-\alpha)^{1-\delta(0)}\alpha^{T-1}, (1-\alpha)^{1-\delta(1)}\alpha^{T-2}, \cdots, (1-\alpha)^{1-\delta((k+1)-1)}\alpha^{T-(k+1)})$.

\textbf{Conclusion:} Since both the base case and the inductive step have been proved as true, by mathematical induction the statement $\beta_{i}=(1-\alpha)^{1-\delta(i-1)} \alpha^{T-i}$ for $i\in\left\{1,2,\cdots,T\right\}$ holds for every positive integer $T$. 
Following Eqn \ref{eqn:Taylor}, the difference between SEAT and the averaged prediction of history networks is controlled by the first order term which reaches 0 only at $\beta_{i}=(1-\alpha)^{1-\delta(i-1)} \alpha^{T-i}$ for $i\in\left\{1,2,\cdots,T\right\}$. Thus, SEAT is hardly be approximate to the averaged prediction of history networks indeed.
\end{proof}
\newpage
\subsection{Proof of Proposition 2}
\begin{prop}
\label{pro:2_re}
(Restated) Assuming that every candidate classifier is updated by SGD-like strategy, meaning $\boldsymbol\theta_{t+1} = \boldsymbol\theta_t -\tau_t \nabla_{\boldsymbol\theta_t}f_{\boldsymbol\theta_t}(x^{\prime}, y)$ with $\tau_1\geq\tau_2\geq\cdots\geq\tau_T>0$, the performance of self-ensemble model depends on learning rate schedules.
\end{prop}
\vspace{-110pt}
\begin{proof}
First we discuss a special case - the change at the $t$-th iteration. Reconsidering the first order term in Eqn \ref{eqn:Taylor}, we have:
\vspace{-20pt}
\begin{equation}
\begin{aligned}
&\sum_{t=1}^{T} (\beta_{t}\boldsymbol\xi^\top) \nabla_{\boldsymbol\xi}f_{\tilde{\boldsymbol\theta}}(x, y) \\
&=\sum_{t=1}^{T} [\beta_{t}(\boldsymbol\theta_t-\tilde{\boldsymbol\theta})] \nabla_{\boldsymbol\xi}f_{\tilde{\boldsymbol\theta}}(x, y) \\
&=\sum_{t=1}^{T} [\beta_{t}((1-(1-\alpha)^{1-\delta(t-1)} \alpha^{T-t})\boldsymbol\theta_t-\tilde{\boldsymbol\theta}_{\mathcal{F}\setminus t})] \nabla_{\boldsymbol\xi}f_{\tilde{\boldsymbol\theta}}(x, y) \\
&=\sum_{t=1}^{T} [\beta_{t}((1-(1-\alpha)^{1-\delta(t-1)} \alpha^{T-t})\boldsymbol\theta_t - (1-(1-\alpha)^{1-\delta(t-1)} \alpha^{T-t})\boldsymbol\theta_{t-1} + \boldsymbol\theta_{t-1} \\
&\quad\quad\quad - (1-\alpha)^{1-\delta(t-1)} \alpha^{T-t}\boldsymbol\theta_{t-1} -\tilde{\boldsymbol\theta}_{\mathcal{F}\setminus t})] \nabla_{\boldsymbol\xi}f_{\tilde{\boldsymbol\theta}}(x, y) \\
&=\sum_{t=1}^{T} [\beta_{t}((1-(1-\alpha)^{1-\delta(t-1)} \alpha^{T-t})\boldsymbol\theta_t - (1-(1-\alpha)^{1-\delta(t-1)} \alpha^{T-t})\boldsymbol\theta_{t-1} + \boldsymbol\theta_{t-1} \\
&\quad\quad\quad - (1-\alpha)^{1-\delta(t-1)} \alpha^{T-t}\boldsymbol\theta_{t-1} - (1-\alpha)^{1-\delta(t-1)} \alpha^{T-t+1}\boldsymbol\theta_{t-1} -\tilde{\boldsymbol\theta}_{\mathcal{F}\setminus t, t-1})] \nabla_{\boldsymbol\xi}f_{\tilde{\boldsymbol\theta}}(x, y) \\
&=\sum_{t=1}^{T} [\beta_{t}((1-(1-\alpha)^{1-\delta(t-1)} \alpha^{T-t})\boldsymbol\theta_t - (1-(1-\alpha)^{1-\delta(t-1)} \alpha^{T-t})\boldsymbol\theta_{t-1} \\
&\quad\quad\quad + (1-(1-\alpha)^{1-\delta(t-1)} \alpha^{T-t} - (1-\alpha)^{1-\delta(t-1)} \alpha^{T-t}\alpha)\boldsymbol\theta_{t-1}-\tilde{\boldsymbol\theta}_{\mathcal{F}\setminus t,t-1})] \nabla_{\boldsymbol\xi}f_{\tilde{\boldsymbol\theta}}(x, y),
\end{aligned}
\end{equation}
so we can deduce by combining:
\begin{equation}
\sum_{t=1}^{T} [\beta_{t}((1-(1-\alpha)^{1-\delta(t-1)} \alpha^{T-t})(\boldsymbol\theta_t - \boldsymbol\theta_{t-1})+ C)] \nabla_{\boldsymbol\xi}f_{\tilde{\boldsymbol\theta}}(x, y),
\end{equation}
where $C=(1-(1-\alpha)^{1-\delta(t-1)} \alpha^{T-t}-(1-\alpha)^{1-\delta(t-1)}\alpha^{T-t}\alpha)\boldsymbol\theta_{t-1}-\tilde{\boldsymbol\theta}_{\mathcal{F}\setminus t,t-1}$.
By using SGD to update $\theta_t$, we have:
\begin{equation}
\sum_{t=1}^{T} [\beta_{t}((1-(1-\alpha)^{1-\delta(t-1)} \alpha^{T-t})(\tau_t \mathbb{E}_{(x, y)}(\nabla_{\boldsymbol\theta_t} \ell(\boldsymbol\theta_t;(x^{\prime}_k, y))+C))]\nabla_{\boldsymbol\xi}f_{\tilde{\boldsymbol\theta}}(x, y).
\end{equation}
Considering $C$ is a constant for the $t$-th update, without changing samples in the $t$-th minibatch, we can conclude that the output of SEAT depends on the learning rate $\tau_t$.


To further analyse the whole training process, we construct $\boldsymbol\theta=\frac{1}{T}\sum_{t=1}^{T}\boldsymbol\theta_t$ and $\tilde{\boldsymbol\xi}=\boldsymbol\theta-\tilde{\boldsymbol\theta}$ to unpack $\tilde{\boldsymbol\theta}_{\mathcal{F}\setminus t}$ for the averaged prediction of history networks, and then reformulate Eqn \ref{eqn:Taylor}:
\begin{equation}
\begin{aligned}
&\sum_{t=1}^{T} (\beta_{t}\tilde{\boldsymbol\xi}^\top) \nabla_{\tilde{\boldsymbol\xi}}f_{\tilde{\boldsymbol\theta}}(x, y) \\
&=\sum_{t=1}^{T} [\beta_{t}(\boldsymbol\theta-\tilde{\boldsymbol\theta})] \nabla_{\tilde{\boldsymbol\xi}}f_{\tilde{\boldsymbol\theta}}(x, y) \\
&=\sum_{t=1}^{T} [\beta_{t}((\frac{1}{T}-(1-\alpha)^{1-\delta(t-1)} \alpha^{T-t})\boldsymbol\theta_t-\tilde{\boldsymbol\theta}_{\mathcal{F}\setminus t})] \nabla_{\tilde{\boldsymbol\xi}}f_{\tilde{\boldsymbol\theta}}(x, y) \\
&=\sum_{t=1}^{T} [\beta_{t}((\frac{1}{T}-(1-\alpha)^{1-\delta(t-1)} \alpha^{T-t})\boldsymbol\theta_t - (\frac{1}{T}-(1-\alpha)^{1-\delta(t-1)} \alpha^{T-t})\boldsymbol\theta_{t-1} + \frac{\boldsymbol\theta_{t-1}}{T} \\
&\quad\quad\quad - (1-\alpha)^{1-\delta(t-1)} \alpha^{T-t}\boldsymbol\theta_{t-1}-\tilde{\boldsymbol\theta}_{\mathcal{F}\setminus t})] \nabla_{\tilde{\boldsymbol\xi}}f_{\tilde{\boldsymbol\theta}}(x, y) \\
&=\sum_{t=1}^{T} [\beta_{t}((\frac{1}{T}-(1-\alpha)^{1-\delta(t-1)} \alpha^{T-t})\boldsymbol\theta_t - (\frac{1}{T}-(1-\alpha)^{1-\delta(t-1)} \alpha^{T-t})\boldsymbol\theta_{t-1} + \frac{\boldsymbol\theta_{t-1}}{T} \\
&\quad\quad\quad - (1-\alpha)^{1-\delta(t-1)} \alpha^{T-t}\boldsymbol\theta_{t-1} - (1-\alpha)^{1-\delta(t-1)} \alpha^{T-t+1}\boldsymbol\theta_{t-1} -\tilde{\boldsymbol\theta}_{\mathcal{F}\setminus t, t-1})] \nabla_{\tilde{\boldsymbol\xi}}f_{\tilde{\boldsymbol\theta}}(x, y) \\
&=\sum_{t=1}^{T} [\beta_{t}((\frac{1}{T}-(1-\alpha)^{1-\delta(t-1)} \alpha^{T-t})\boldsymbol\theta_t - (\frac{1}{T}-(1-\alpha)^{1-\delta(t-1)} \alpha^{T-t})\boldsymbol\theta_{t-1} + (\frac{1}{T} \\
&\quad\quad\quad -(1-\alpha)^{1-\delta(t-1)} \alpha^{T-t} - (1-\alpha)^{1-\delta(t-1)} \alpha^{T-t}\alpha)\boldsymbol\theta_{t-1}-\tilde{\boldsymbol\theta}_{\mathcal{F}\setminus t,t-1})] \nabla_{\tilde{\boldsymbol\xi}}f_{\tilde{\boldsymbol\theta}}(x, y) \\
&=\sum_{t=1}^{T} [\beta_{t}((\frac{1}{T}-(1-\alpha)^{1-\delta(t-1)} \alpha^{T-t})(\boldsymbol\theta_t - \boldsymbol\theta_{t-1})+ C^{\prime})] \nabla_{\boldsymbol\xi}f_{\tilde{\boldsymbol\theta}}(x, y),
\end{aligned}
\end{equation}
where $C^{\prime}=(\frac{1}{T}-(1-\alpha)^{1-\delta(t-1)} \alpha^{T-t}-(1-\alpha)^{1-\delta(t-1)}\alpha^{T-t}\alpha)\boldsymbol\theta_{t-1}-\tilde{\boldsymbol\theta}_{\mathcal{F}\setminus t,t-1}$. Likewise, the above equation can be further deduced:
\begin{equation}
\sum_{t=1}^{T} [\beta_{t}((\frac{1}{T}-(1-\alpha)^{1-\delta(t-1)} \alpha^{T-t})(\tau_t \mathbb{E}_{(x, y)}(\nabla_{\boldsymbol\theta_t} \ell(\boldsymbol\theta_t;(x^{\prime}_k, y))+C^{\prime}))]\nabla_{\boldsymbol\xi}f_{\tilde{\boldsymbol\theta}}(x, y),
\end{equation}
which means the difference between $\bar{f}_{\mathcal{F}}(x,y)$ and $f_{\tilde{\boldsymbol\theta}}(x, y)$ depends on learning rate schedules.
\end{proof}

\end{document}

%% file: math_commands.tex
\usepackage{amsmath,amsfonts,bm}

\def\eqref#1{equation~\ref{#1}}

\def\1{\bm{1}}

\DeclareMathAlphabet{\mathsfit}{\encodingdefault}{\sfdefault}{m}{sl}
\SetMathAlphabet{\mathsfit}{bold}{\encodingdefault}{\sfdefault}{bx}{n}













%% file: iclr2022_conference.bbl
\begin{thebibliography}{54}
\providecommand{\natexlab}[1]{#1}
\providecommand{\url}[1]{\texttt{#1}}
\expandafter\ifx\csname urlstyle\endcsname\relax
  \providecommand{\doi}[1]{doi: #1}\else
  \providecommand{\doi}{doi: \begingroup \urlstyle{rm}\Url}\fi

\bibitem[Andriushchenko et~al.(2020)Andriushchenko, Croce, Flammarion, and
  Hein]{ACFH2020square}
Maksym Andriushchenko, Francesco Croce, Nicolas Flammarion, and Matthias Hein.
\newblock Square attack: a query-efficient black-box adversarial attack via
  random search.
\newblock In \emph{ECCV}, 2020.

\bibitem[Bai et~al.(2019)Bai, Feng, Wang, Dai, Xia, and Jiang]{bai2019hilbert}
Yang Bai, Yan Feng, Yisen Wang, Tao Dai, Shu-Tao Xia, and Yong Jiang.
\newblock Hilbert-based generative defense for adversarial examples.
\newblock In \emph{ICCV}, 2019.

\bibitem[Bai et~al.(2021)Bai, Zeng, Jiang, Xia, Ma, and Wang]{bai2021improving}
Yang Bai, Yuyuan Zeng, Yong Jiang, Shu-Tao Xia, Xingjun Ma, and Yisen Wang.
\newblock Improving adversarial robustness via channel-wise activation
  suppressing.
\newblock In \emph{ICLR}, 2021.

\bibitem[Baluja \& Fischer(2017)Baluja and
  Fischer]{DBLP:journals/corr/BalujaF17}
Shumeet Baluja and Ian Fischer.
\newblock Adversarial transformation networks: Learning to generate adversarial
  examples.
\newblock \emph{arXiv preprint arXiv:1703.09387}, 2017.

\bibitem[Bartlett \& Mendelson(2001)Bartlett and
  Mendelson]{DBLP:conf/colt/BartlettM01}
Peter~L. Bartlett and Shahar Mendelson.
\newblock Rademacher and gaussian complexities: Risk bounds and structural
  results.
\newblock In \emph{COLT}, 2001.

\bibitem[Blanchet \& Murthy(2019)Blanchet and
  Murthy]{DBLP:journals/mor/BlanchetM19}
Jose~H. Blanchet and Karthyek R.~A. Murthy.
\newblock Quantifying distributional model risk via optimal transport.
\newblock \emph{Math. Oper. Res.}, 2019.

\bibitem[Brock et~al.(2019)Brock, Donahue, and
  Simonyan]{DBLP:conf/iclr/BrockDS19}
Andrew Brock, Jeff Donahue, and Karen Simonyan.
\newblock Large scale {GAN} training for high fidelity natural image synthesis.
\newblock In \emph{ICLR}, 2019.

\bibitem[Carlini \& Wagner(2017)Carlini and Wagner]{DBLP:conf/sp/Carlini017}
Nicholas Carlini and David~A. Wagner.
\newblock Towards evaluating the robustness of neural networks.
\newblock In \emph{S\&P}, 2017.

\bibitem[Caruana et~al.(2004)Caruana, Niculescu{-}Mizil, Crew, and
  Ksikes]{DBLP:conf/icml/CaruanaNCK04}
Rich Caruana, Alexandru Niculescu{-}Mizil, Geoff Crew, and Alex Ksikes.
\newblock Ensemble selection from libraries of models.
\newblock In \emph{ICML}, 2004.

\bibitem[Child(2021)]{DBLP:conf/iclr/Child21}
Rewon Child.
\newblock Very deep vaes generalize autoregressive models and can outperform
  them on images.
\newblock In \emph{ICLR}, 2021.

\bibitem[Croce \& Hein(2020)Croce and Hein]{DBLP:conf/icml/Croce020a}
Francesco Croce and Matthias Hein.
\newblock Reliable evaluation of adversarial robustness with an ensemble of
  diverse parameter-free attacks.
\newblock In \emph{ICML}, 2020.

\bibitem[Devlin et~al.(2019)Devlin, Chang, Lee, and
  Toutanova]{DBLP:conf/naacl/DevlinCLT19}
Jacob Devlin, Ming{-}Wei Chang, Kenton Lee, and Kristina Toutanova.
\newblock {BERT:} pre-training of deep bidirectional transformers for language
  understanding.
\newblock In \emph{NAACL}, 2019.

\bibitem[Ding et~al.(2020)Ding, Sharma, Lui, and
  Huang]{DBLP:conf/iclr/DingSLH20}
Gavin~Weiguang Ding, Yash Sharma, Kry Yik~Chau Lui, and Ruitong Huang.
\newblock {MMA} training: Direct input space margin maximization through
  adversarial training.
\newblock In \emph{ICLR}, 2020.

\bibitem[Dong et~al.(2018)Dong, Liao, Pang, Su, Zhu, Hu, and
  Li]{DBLP:conf/cvpr/DongLPS0HL18}
Yinpeng Dong, Fangzhou Liao, Tianyu Pang, Hang Su, Jun Zhu, Xiaolin Hu, and
  Jianguo Li.
\newblock Boosting adversarial attacks with momentum.
\newblock In \emph{CVPR}, 2018.

\bibitem[Garipov et~al.(2018)Garipov, Izmailov, Podoprikhin, Vetrov, and
  Wilson]{DBLP:conf/nips/GaripovIPVW18}
Timur Garipov, Pavel Izmailov, Dmitrii Podoprikhin, Dmitry~P. Vetrov, and
  Andrew~Gordon Wilson.
\newblock Loss surfaces, mode connectivity, and fast ensembling of dnns.
\newblock In \emph{NeurIPS}, 2018.

\bibitem[Goodfellow \& Vinyals(2015)Goodfellow and
  Vinyals]{DBLP:journals/corr/GoodfellowV14}
Ian~J. Goodfellow and Oriol Vinyals.
\newblock Qualitatively characterizing neural network optimization problems.
\newblock In \emph{ICLR}, 2015.

\bibitem[Goodfellow et~al.(2015)Goodfellow, Shlens, and
  Szegedy]{DBLP:journals/corr/GoodfellowSS14}
Ian~J. Goodfellow, Jonathon Shlens, and Christian Szegedy.
\newblock Explaining and harnessing adversarial examples.
\newblock In \emph{ICLR}, 2015.

\bibitem[Gowal et~al.(2020)Gowal, Qin, Uesato, Mann, and
  Kohli]{gowal2020uncovering}
Sven Gowal, Chongli Qin, Jonathan Uesato, Timothy Mann, and Pushmeet Kohli.
\newblock Uncovering the limits of adversarial training against norm-bounded
  adversarial examples.
\newblock \emph{arXiv preprint arXiv:2010.03593}, 2020.

\bibitem[He et~al.(2016)He, Zhang, Ren, and Sun]{DBLP:conf/cvpr/HeZRS16}
Kaiming He, Xiangyu Zhang, Shaoqing Ren, and Jian Sun.
\newblock Deep residual learning for image recognition.
\newblock In \emph{CVPR}, 2016.

\bibitem[Hendrycks \& Gimpel(2016)Hendrycks and
  Gimpel]{DBLP:journals/corr/HendrycksG16}
Dan Hendrycks and Kevin Gimpel.
\newblock Bridging nonlinearities and stochastic regularizers with gaussian
  error linear units.
\newblock \emph{arXiv preprint arXiv:1606.08415}, 2016.

\bibitem[Ho et~al.(2020)Ho, Jain, and Abbeel]{DBLP:conf/nips/HoJA20}
Jonathan Ho, Ajay Jain, and Pieter Abbeel.
\newblock Denoising diffusion probabilistic models.
\newblock In \emph{NeurIPS}, 2020.

\bibitem[Huang et~al.(2021)Huang, Wang, Erfani, Gu, Bailey, and
  Ma]{huang2021exploring}
Hanxun Huang, Yisen Wang, Sarah~Monazam Erfani, Quanquan Gu, James Bailey, and
  Xingjun Ma.
\newblock Exploring architectural ingredients of adversarially robust deep
  neural networks.
\newblock In \emph{NeurIPS}, 2021.

\bibitem[Jumper et~al.(2021)Jumper, Evans, Pritzel, Green, Figurnov,
  Ronneberger, Tunyasuvunakool, Bates, {\v{Z}}{\'\i}dek, Potapenko, Bridgland,
  Meyer, Kohl, Ballard, Cowie, Romera-Paredes, Nikolov, Jain, Adler, Back,
  Petersen, Reiman, Clancy, Zielinski, Steinegger, Pacholska, Berghammer,
  Bodenstein, Silver, Vinyals, Senior, Kavukcuoglu, Kohli, and
  Hassabis]{AlphaFold2021}
John Jumper, Richard Evans, Alexander Pritzel, Tim Green, Michael Figurnov,
  Olaf Ronneberger, Kathryn Tunyasuvunakool, Russ Bates, Augustin
  {\v{Z}}{\'\i}dek, Anna Potapenko, Alex Bridgland, Clemens Meyer, Simon A~A
  Kohl, Andrew~J Ballard, Andrew Cowie, Bernardino Romera-Paredes, Stanislav
  Nikolov, Rishub Jain, Jonas Adler, Trevor Back, Stig Petersen, David Reiman,
  Ellen Clancy, Michal Zielinski, Martin Steinegger, Michalina Pacholska, Tamas
  Berghammer, Sebastian Bodenstein, David Silver, Oriol Vinyals, Andrew~W
  Senior, Koray Kavukcuoglu, Pushmeet Kohli, and Demis Hassabis.
\newblock Highly accurate protein structure prediction with {AlphaFold}.
\newblock \emph{Nature}, 2021.

\bibitem[Kannan et~al.(2018)Kannan, Kurakin, and
  Goodfellow]{DBLP:journals/corr/abs-1803-06373}
Harini Kannan, Alexey Kurakin, and Ian~J. Goodfellow.
\newblock Adversarial logit pairing.
\newblock \emph{arXiv preprint arXiv:1803.06373}, 2018.

\bibitem[Krizhevsky et~al.(2012)Krizhevsky, Sutskever, and
  Hinton]{DBLP:conf/nips/KrizhevskySH12}
Alex Krizhevsky, Ilya Sutskever, and Geoffrey~E. Hinton.
\newblock Imagenet classification with deep convolutional neural networks.
\newblock In \emph{NeurIPS}, 2012.

\bibitem[Li et~al.(2018)Li, Xu, Taylor, Studer, and Goldstein]{visualloss}
Hao Li, Zheng Xu, Gavin Taylor, Christoph Studer, and Tom Goldstein.
\newblock Visualizing the loss landscape of neural nets.
\newblock In \emph{NeurIPS}, 2018.

\bibitem[Liao et~al.(2018)Liao, Liang, Dong, Pang, Hu, and
  Zhu]{DBLP:conf/cvpr/LiaoLDPH018}
Fangzhou Liao, Ming Liang, Yinpeng Dong, Tianyu Pang, Xiaolin Hu, and Jun Zhu.
\newblock Defense against adversarial attacks using high-level representation
  guided denoiser.
\newblock In \emph{CVPR}, 2018.

\bibitem[Ma et~al.(2020)Ma, Niu, Gu, Wang, Zhao, Bailey, and
  Lu]{ma2019understanding}
Xingjun Ma, Yuhao Niu, Lin Gu, Yisen Wang, Yitian Zhao, James Bailey, and Feng
  Lu.
\newblock Understanding adversarial attacks on deep learning based medical
  image analysis systems.
\newblock \emph{Pattern Recognition}, 2020.

\bibitem[Madry et~al.(2018)Madry, Makelov, Schmidt, Tsipras, and
  Vladu]{DBLP:conf/iclr/MadryMSTV18}
Aleksander Madry, Aleksandar Makelov, Ludwig Schmidt, Dimitris Tsipras, and
  Adrian Vladu.
\newblock Towards deep learning models resistant to adversarial attacks.
\newblock In \emph{ICLR}, 2018.

\bibitem[Miyato et~al.(2019)Miyato, Maeda, Koyama, and
  Ishii]{DBLP:journals/pami/MiyatoMKI19}
Takeru Miyato, Shin{-}ichi Maeda, Masanori Koyama, and Shin Ishii.
\newblock Virtual adversarial training: {A} regularization method for
  supervised and semi-supervised learning.
\newblock \emph{{IEEE} Trans. Pattern Anal. Mach. Intell.}, 2019.

\bibitem[Niu et~al.(2021)Niu, Guo, and Wang]{niu2021moire}
Dantong Niu, Ruohao Guo, and Yisen Wang.
\newblock Moiré attack (ma): A new potential risk of screen photos.
\newblock In \emph{NeurIPS}, 2021.

\bibitem[Pang et~al.(2019)Pang, Xu, Du, Chen, and
  Zhu]{DBLP:conf/icml/PangXDCZ19}
Tianyu Pang, Kun Xu, Chao Du, Ning Chen, and Jun Zhu.
\newblock Improving adversarial robustness via promoting ensemble diversity.
\newblock In \emph{ICML}, 2019.

\bibitem[Papernot et~al.(2016)Papernot, McDaniel, Wu, Jha, and
  Swami]{DBLP:conf/sp/PapernotM0JS16}
Nicolas Papernot, Patrick~D. McDaniel, Xi~Wu, Somesh Jha, and Ananthram Swami.
\newblock Distillation as a defense to adversarial perturbations against deep
  neural networks.
\newblock In \emph{S\&P}, 2016.

\bibitem[Qin et~al.(2020)Qin, Frosst, Sabour, Raffel, Cottrell, and
  Hinton]{DBLP:conf/iclr/QinFSRCH20}
Yao Qin, Nicholas Frosst, Sara Sabour, Colin Raffel, Garrison~W. Cottrell, and
  Geoffrey~E. Hinton.
\newblock Detecting and diagnosing adversarial images with class-conditional
  capsule reconstructions.
\newblock In \emph{ICLR}, 2020.

\bibitem[Rebuffi et~al.(2021)Rebuffi, Gowal, Calian, Stimberg, Wiles, and
  Mann]{rebuffi2021fixing}
Sylvestre-Alvise Rebuffi, Sven Gowal, Dan~A Calian, Florian Stimberg, Olivia
  Wiles, and Timothy Mann.
\newblock Fixing data augmentation to improve adversarial robustness.
\newblock \emph{arXiv preprint arXiv:2103.01946}, 2021.

\bibitem[Rice et~al.(2020)Rice, Wong, and Kolter]{DBLP:conf/icml/RiceWK20}
Leslie Rice, Eric Wong, and J.~Zico Kolter.
\newblock Overfitting in adversarially robust deep learning.
\newblock In \emph{ICML}, 2020.

\bibitem[Ross \& Doshi{-}Velez(2018)Ross and
  Doshi{-}Velez]{DBLP:conf/aaai/RossD18}
Andrew~Slavin Ross and Finale Doshi{-}Velez.
\newblock Improving the adversarial robustness and interpretability of deep
  neural networks by regularizing their input gradients.
\newblock In \emph{AAAI}, 2018.

\bibitem[Sak et~al.(2015)Sak, Senior, Rao, and
  Beaufays]{DBLP:conf/interspeech/SakSRB15}
Hasim Sak, Andrew~W. Senior, Kanishka Rao, and Fran{\c{c}}oise Beaufays.
\newblock Fast and accurate recurrent neural network acoustic models for speech
  recognition.
\newblock In \emph{INTERSPEECH}, 2015.

\bibitem[Si \& Yan(2021)Si and Yan]{AlphaFold2020}
Yunda Si and Chengfei Yan.
\newblock Improved protein contact prediction using dimensional hybrid residual
  networks and singularity enhanced loss function.
\newblock \emph{bioRxiv}, 2021.

\bibitem[Szegedy et~al.(2014)Szegedy, Zaremba, Sutskever, Bruna, Erhan,
  Goodfellow, and Fergus]{DBLP:journals/corr/SzegedyZSBEGF13}
Christian Szegedy, Wojciech Zaremba, Ilya Sutskever, Joan Bruna, Dumitru Erhan,
  Ian~J. Goodfellow, and Rob Fergus.
\newblock Intriguing properties of neural networks.
\newblock In \emph{ICLR}, 2014.

\bibitem[Tram{\`{e}}r et~al.(2018)Tram{\`{e}}r, Kurakin, Papernot, Goodfellow,
  Boneh, and McDaniel]{DBLP:conf/iclr/TramerKPGBM18}
Florian Tram{\`{e}}r, Alexey Kurakin, Nicolas Papernot, Ian~J. Goodfellow, Dan
  Boneh, and Patrick~D. McDaniel.
\newblock Ensemble adversarial training: Attacks and defenses.
\newblock In \emph{ICLR}, 2018.

\bibitem[Vaswani et~al.(2017)Vaswani, Shazeer, Parmar, Uszkoreit, Jones, Gomez,
  Kaiser, and Polosukhin]{DBLP:conf/nips/VaswaniSPUJGKP17}
Ashish Vaswani, Noam Shazeer, Niki Parmar, Jakob Uszkoreit, Llion Jones,
  Aidan~N. Gomez, Lukasz Kaiser, and Illia Polosukhin.
\newblock Attention is all you need.
\newblock In \emph{NeurIPS}, 2017.

\bibitem[Villani(2003)]{Villani2003TopicsIO}
Cedric Villani.
\newblock Topics in optimal transportation.(books).
\newblock \emph{OR/MS Today}, 2003.

\bibitem[Wang et~al.(2020{\natexlab{a}})Wang, Li, Liu, and
  Lin]{wang2020hamiltonian}
Hongjun Wang, Guanbin Li, Xiaobai Liu, and Liang Lin.
\newblock A hamiltonian monte carlo method for probabilistic adversarial attack
  and learning.
\newblock \emph{IEEE Transactions on Pattern Analysis and Machine
  Intelligence}, 2020{\natexlab{a}}.

\bibitem[Wang et~al.(2020{\natexlab{b}})Wang, Wang, Li, Zhang, and
  Lin]{DBLP:conf/cvpr/0005WLZL20}
Hongjun Wang, Guangrun Wang, Ya~Li, Dongyu Zhang, and Liang Lin.
\newblock Transferable, controllable, and inconspicuous adversarial attacks on
  person re-identification with deep mis-ranking.
\newblock In \emph{CVPR}, 2020{\natexlab{b}}.

\bibitem[Wang et~al.(2017)Wang, Deng, Pu, and Huang]{wang2017residual}
Yisen Wang, Xuejiao Deng, Songbai Pu, and Zhiheng Huang.
\newblock Residual convolutional ctc networks for automatic speech recognition.
\newblock \emph{arXiv preprint arXiv:1702.07793}, 2017.

\bibitem[Wang et~al.(2019)Wang, Ma, Bailey, Yi, Zhou, and Gu]{wang2019dynamic}
Yisen Wang, Xingjun Ma, James Bailey, Jinfeng Yi, Bowen Zhou, and Quanquan Gu.
\newblock On the convergence and robustness of adversarial training.
\newblock In \emph{ICML}, 2019.

\bibitem[Wang et~al.(2020{\natexlab{c}})Wang, Zou, Yi, Bailey, Ma, and
  Gu]{DBLP:conf/iclr/0001ZY0MG20}
Yisen Wang, Difan Zou, Jinfeng Yi, James Bailey, Xingjun Ma, and Quanquan Gu.
\newblock Improving adversarial robustness requires revisiting misclassified
  examples.
\newblock In \emph{ICLR}, 2020{\natexlab{c}}.

\bibitem[Wu et~al.(2020)Wu, Xia, and Wang]{wu2020adversarial}
Dongxian Wu, Shu-Tao Xia, and Yisen Wang.
\newblock Adversarial weight perturbation helps robust generalization.
\newblock In \emph{NeurIPS}, 2020.

\bibitem[Xu et~al.(2018)Xu, Evans, and Qi]{DBLP:conf/ndss/Xu0Q18}
Weilin Xu, David Evans, and Yanjun Qi.
\newblock Feature squeezing: Detecting adversarial examples in deep neural
  networks.
\newblock In \emph{NDSS}, 2018.

\bibitem[Zagoruyko \& Komodakis(2016)Zagoruyko and
  Komodakis]{DBLP:journals/corr/ZagoruykoK16}
Sergey Zagoruyko and Nikos Komodakis.
\newblock Wide residual networks.
\newblock \emph{arXiv preprint arXiv:1605.07146}, 2016.

\bibitem[Zhang et~al.(2019)Zhang, Yu, Jiao, Xing, Ghaoui, and
  Jordan]{DBLP:conf/icml/ZhangYJXGJ19}
Hongyang Zhang, Yaodong Yu, Jiantao Jiao, Eric~P. Xing, Laurent~El Ghaoui, and
  Michael~I. Jordan.
\newblock Theoretically principled trade-off between robustness and accuracy.
\newblock In \emph{ICML}, 2019.

\bibitem[Zhang et~al.(2020)Zhang, Xu, Han, Niu, Cui, Sugiyama, and
  Kankanhalli]{DBLP:conf/icml/ZhangXH0CSK20}
Jingfeng Zhang, Xilie Xu, Bo~Han, Gang Niu, Lizhen Cui, Masashi Sugiyama, and
  Mohan~S. Kankanhalli.
\newblock Attacks which do not kill training make adversarial learning
  stronger.
\newblock In \emph{ICML}, 2020.

\bibitem[Zhang et~al.(2021)Zhang, Zhu, Niu, Han, Sugiyama, and
  Kankanhalli]{DBLP:conf/iclr/ZhangZ00SK21}
Jingfeng Zhang, Jianing Zhu, Gang Niu, Bo~Han, Masashi Sugiyama, and Mohan~S.
  Kankanhalli.
\newblock Geometry-aware instance-reweighted adversarial training.
\newblock In \emph{ICLR}, 2021.

\end{thebibliography}
